\documentclass{article} %
\usepackage{iclr2027_conference,times}

\usepackage{amsmath,amsfonts,bm}

\def\eqref#1{equation~\ref{#1}}

\def\1{\bm{1}}

\DeclareMathAlphabet{\mathsfit}{\encodingdefault}{\sfdefault}{m}{sl}
\SetMathAlphabet{\mathsfit}{bold}{\encodingdefault}{\sfdefault}{bx}{n}

\usepackage[dvipsnames]{xcolor}
\usepackage[pagebackref,breaklinks,colorlinks,allcolors=Plum]{hyperref}

\usepackage{url}

\usepackage{booktabs}
\usepackage{multirow}
\usepackage{makecell}
\usepackage{array}
\usepackage[table]{xcolor}
\usepackage{tabularx}
\usepackage{graphicx}
\usepackage{subcaption}
\usepackage{wrapfig}
\usepackage{pifont}
\usepackage{color}
\usepackage{amssymb}
\usepackage{placeins}
\usepackage{needspace}
\usepackage{enumitem}

\usepackage[capitalize,noabbrev]{cleveref}

\definecolor{backbonegray}{RGB}{248,248,248}

\definecolor{avggray}{RGB}{236,236,236}

\newcommand{\paragrapht}[1]{\noindent\textbf{#1}}

\definecolor{backboneavggray}{RGB}{236,236,236}

\definecolor{ourslavender}{RGB}{242,239,250}

\definecolor{redavggray}{RGB}{228,225,238}

\newcommand{\venue}[1]{%
  \hspace{0.18em}{%
    \color{venuegray}%
  }%
}
\newcommand{\ie}{\textit{i.e.}}
\newcommand{\eg}{\textit{e.g.}}
\newcommand{\cmark}{\ding{51}}
\newcommand{\xmark}{\ding{55}}

\usepackage{xspace}
\newcommand{\narrowtexttt}[1]{\scalebox{0.9}[0.98]{\texttt{#1}}}

\newcommand\ours{\narrowtexttt{\textbf{RED}}\xspace}

\newcolumntype{C}[1]{>{\centering\arraybackslash}p{#1}}
\newcolumntype{M}[1]{>{\raggedright\arraybackslash}p{#1}}
\newcolumntype{Y}{>{\centering\arraybackslash\hspace{0pt}}X}
\newcolumntype{P}{>{\centering\arraybackslash}p{0.18cm}}

\usepackage{lipsum}

\title{Relevant Evidence Decoding for Audio-Visual Hallucination Mitigation}

\author{
Hyunjae Ra$^{1}$ \quad
Aecheon Jung$^{1}$ \quad
Jungin Park$^{2}$ \quad
Sungeun Hong$^{1}$\thanks{Corresponding author.} \\
$^{1}$Sungkyunkwan University
\qquad
$^{2}$Daegu Gyeongbuk Institute of Science and Technology (DGIST)
}

\iclrfinalcopy %
\begin{document}

\maketitle
\lhead{} 
\renewcommand{\headrulewidth}{0pt}
\vspace{-2mm}

\begin{abstract}
Audio-Visual Large Language Models (AV-LLMs) remain prone to cross-modal hallucinations, where one modality incorrectly affects predictions about another. Although contrastive decoding reduces hallucinations in vision-language models, its direct extension to AV-LLMs overlooks a key challenge: different questions require different perceptual evidence, including audio, video, or their interaction.
Notably, we observe that joint audio-visual inference can weaken the prediction even when a model can recover the
correct answer from a single informative modality.
For example, when asked which instrument is heard, a model may correctly predict violin from the audio alone. Once a video showing a guitar is added, its confidence in violin may drop.
In this paper, we introduce \textbf{R}elevant \textbf{E}vidence \textbf{D}ecoding (\ours), a training-free method that identifies question-relevant evidence and selectively strengthens its contribution.
\ours uses pointwise mutual information to quantify the predictive support provided by audio and video beyond the question alone. It decomposes their joint contribution into audio, video, and residual interaction components. 
A question-only inference pass determines the required evidence type, after which the model augments the original audio-visual prediction with the corresponding PMI contribution.
Across three audio-visual hallucination benchmarks and three AV-LLMs, \ours improves accuracy over standard decoding by up to 7\% on CMM, 6.3\% on AVHBench, and 3.8\% on SVHalluc, with an average relative time to first token of 1.5$\times$ standard decoding.

\end{abstract}

\vspace{-2mm}

\section{Introduction}
\vspace{-2mm}

Multimodal large language models (MLLMs) extend LLMs to perceptual inputs such as vision~\citep{LLaVA_2023_NeurIPS,Flamingo_2022_NeurIPS} and audio~\citep{VideoLLaMA2_2024_arXiv,Qwen2_5_Omni_2025_arXiv}, but additional modalities also introduce new sources of hallucination~\citep{VidHalluc_2025_CVPR,HalluAudio_2026_ACL}. Prior work has linked these hallucinations to modality dominance~\citep{CMM_2025_NeurIPS,Ye_2026_AAAI}, cross-modal interference~\citep{Li_2026_SFFL,Cai_2025_ModalityInterference}, and spurious inter-modality correlations~\citep{CMM_2025_NeurIPS}. Consequently, more perceptual information does not necessarily yield more reliable predictions. In particular, we find that joint audio-visual inference can weaken a prediction even when the modality required by the question alone provides sufficient evidence for the correct answer.

Contrastive decoding (CD) has emerged as a promising approach to mitigate hallucinations in LLMs~\citep{CAD_2024_NAACL,DoLa_2024_ICLR} and VLMs~\citep{VCD_2024_CVPR,ICD_2024_FindingsACL,CODE_2024_NeurIPS}. CD compares predictions from the same model under different conditioning inputs and uses their difference to adjust the original prediction without additional training. Recent methods have extended this idea to AV-LLMs by accounting for modality dominance~\citep{AVCD_2025_NeurIPS}, adaptively weighting modality-specific branches~\citep{MAD_2026_CVPR}, or modeling cross-modal information flow~\citep{ASD_2026_ICML}. However, these methods focus on balancing or integrating multiple modalities rather than aligning the decoding correction with the evidence required by each question.

\begin{figure*}[t]
    \centering
    \includegraphics[width=0.95\textwidth]{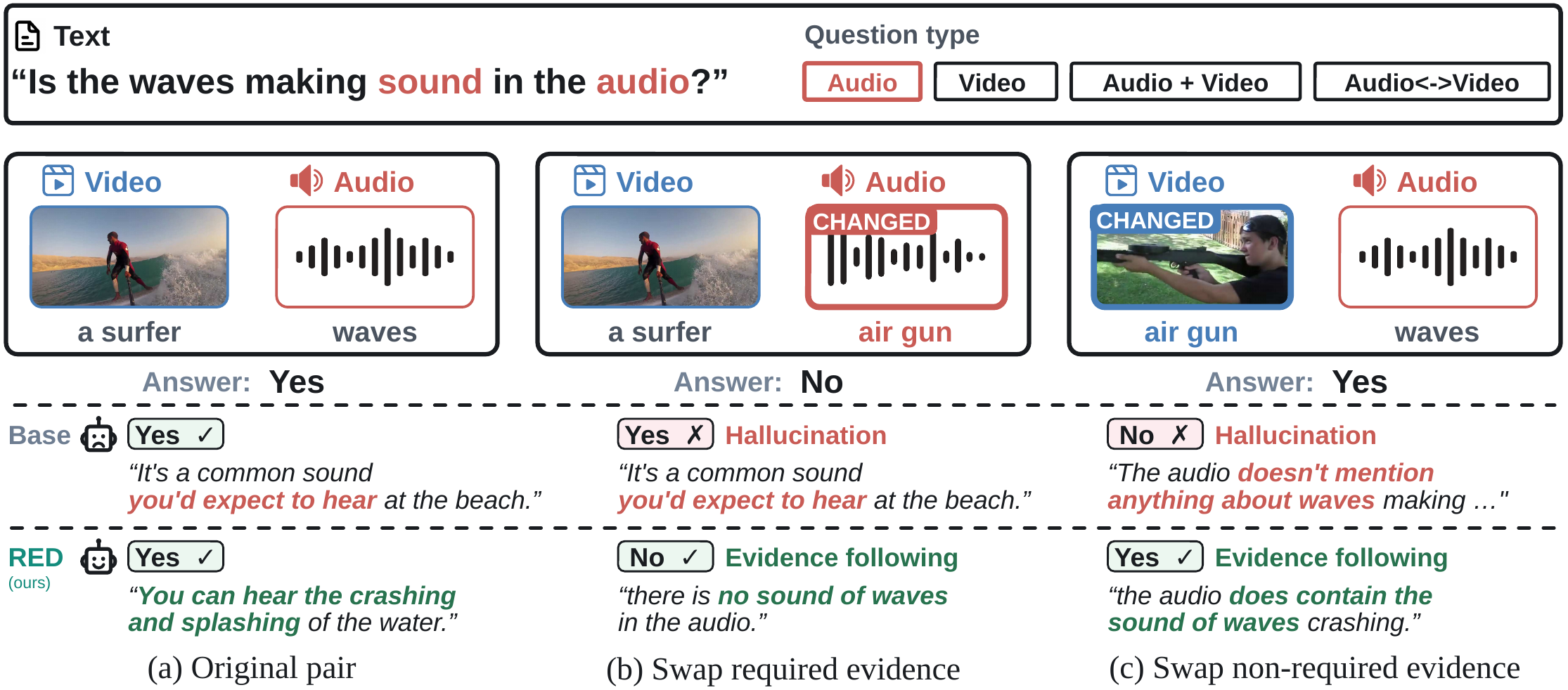}
    \caption{\textbf{Motivation of this study.}
    (a) A correct prediction does not necessarily indicate successful hallucination mitigation.
    (b) When question-required evidence changes, the prediction should change accordingly.
    (c) When non-required evidence changes, it should remain unchanged.
    The base model violates both behaviors, whereas \ours follows the evidence required by the question.
    }
    \label{fig:motivation}
    \vspace{-1.2em}
\end{figure*}

In practice, a correct prediction does not necessarily imply appropriate use of perceptual evidence~\citep{Xiao_2024_CVPR,MIRAGEPerf_2025_NeurIPS}.
A model may produce the correct answer from language priors~\citep{M3ID_2024_CVPR}, correlated cues, or perceptual evidence that is not required by the question~\citep{ACPO_2026_ECCV}.
For example, an audio-dependent question can be answered correctly using correlated visual evidence while the relevant audio evidence has little effect on the prediction, as shown in \cref{fig:motivation}(a).
We argue that the required evidence should instead have a direct and selective influence on prediction: \textit{changing question-required evidence should alter the prediction, whereas changing irrelevant evidence should have substantially less effect} (please see \cref{fig:motivation}(b) and (c)).
The central challenge is to couple two distinct quantities: what evidence the question requires and how strongly that evidence supports each candidate prediction. This motivates evidence-following decoding, which aligns the correction with question-required predictive support rather than with any signal that happens to favor the answer.

In this study, we introduce \textbf{R}elevant \textbf{E}vidence \textbf{D}ecoding \textbf{(\ours)}, a training-free decoding method that identifies the evidence required by the question and aligns the decoding correction with the measured predictive support of that evidence. A question-only forward pass identifies the required evidence. Pointwise Mutual Information (PMI)~\citep{Church_Hanks_1990_CL} measures its predictive support beyond the question and language context by decomposing the joint audio-visual contribution into audio, video, and residual interaction components. \ours refines the original audio-visual prediction using the PMI contribution corresponding to the required evidence.
Contrary to existing CD methods that modulate modality-level branches~\citep{AVCD_2025_NeurIPS,MAD_2026_CVPR} or amplify the overall contribution of additional conditioning information~\citep{CAD_2024_NAACL,M3ID_2024_CVPR}, \ours directly encourages predictions to follow question-required evidence rather than question-irrelevant cues.

We evaluate our method with three backbones (\eg, VideoLLaMA2.1-7B-AV~\citep{VideoLLaMA2_2024_arXiv}, Qwen2.5-Omni-3B, and -7B~\citep{Qwen2_5_Omni_2025_arXiv}) on three audio-visual hallucination benchmarks: CMM~\citep{CMM_2025_NeurIPS}, AVHBench~\citep{AVHBench_2025_ICLR}, and SVHalluc~\citep{SVHalluc_2026_CVPR}. 
\ours consistently outperforms existing decoding methods across all benchmarks and backbones. Results on two general audio-visual question answering benchmarks~\citep{MUSIC_AVQA_2022_CVPR,Kim_2026_Interspeech} further show that its benefits transfer to general audio-visual reasoning. Extensive analyses show complementary gains from evidence selection and PMI decomposition, with RED's average relative time to first token (TTFT) at 1.5$\times$ standard decoding.

\begin{itemize}[leftmargin=*, itemsep=1pt]
\item We characterize evidence following through controlled modality interventions, showing that even correct AV-LLM predictions can respond weakly to question-required evidence and remain sensitive to non-required evidence.
\item We propose \ours, a training-free CD method that uses PMI to decompose audio-visual predictive changes relative to a common question-only reference and selectively strengthens the contribution corresponding to question-required evidence.
   \item Across three AV-LLMs and five benchmarks, \ours generally outperforms existing decoding methods on both hallucination and general audio-visual reasoning tasks, while extensive analyses validate the effectiveness of evidence selection and PMI decomposition.
\end{itemize}

\section{Related Work}

\paragrapht{Multimodal hallucination.}
Hallucination in vision-language models (VLMs) has been widely studied, particularly in relation to object hallucination and excessive reliance on language priors~\citep{POPE_2023_EMNLP,HallusionBench_2024_CVPR}.
When models jointly process audio and video, additional failures can arise from modality dominance, cross-modal interference, and spurious inter-modality correlations~\citep{AVHBench_2025_ICLR,CMM_2025_NeurIPS,SVHalluc_2026_CVPR}.
Controlled perturbations and modality interventions reveal how predictions depend on individual input sources~\citep{MIRAGEPerf_2025_NeurIPS,ACPO_2026_ECCV}.
However, input dependence alone does not indicate whether a prediction follows the evidence required by a particular question.
Recent training-based work also promotes sensitivity to relevant modalities and invariance to irrelevant ones through modality-aware preference optimization~\citep{MoDDPO_2026_CVPR}.
We study \emph{evidence following} as a question-conditioned criterion that distinguishes what evidence is required from how it contributes to the prediction: predictions should respond to required evidence while remaining stable to non-required evidence.

\paragrapht{Decoding-based hallucination mitigation.}
Contrastive decoding (CD) mitigates hallucination by adjusting predictions using differences between distributions obtained under alternative input conditions~\citep{VCD_2024_CVPR,ICD_2024_FindingsACL,CODE_2024_NeurIPS,SID_2025_ICLR}.
Existing VLM methods construct these contrasts from distorted visual inputs, perturbed instructions, generated descriptions, or self-introspective signals~\citep{VCD_2024_CVPR,ICD_2024_FindingsACL,CODE_2024_NeurIPS,SID_2025_ICLR}.
For audio-visual LLMs, AVCD~\citep{AVCD_2025_NeurIPS} constructs modality-dependent contrasts, MAD~\citep{MAD_2026_CVPR} adaptively combines modality-conditioned branches, and ASD~\citep{ASD_2026_ICML} promotes cross-modal information flow.
Information-theoretic decoding methods emphasize input-dependent information through context-aware, mutual-information, or PMI-based objectives~\citep{CAD_2024_NAACL,M3ID_2024_CVPR,DynamicPMI_2026_FindingsACL}.
Among these methods, MAD~\citep{MAD_2026_CVPR} is most closely related to \ours because both account for question-dependent evidence requirements.
While MAD adaptively weights modality-conditioned branches, \ours selects among contributions defined relative to a common question-only reference.
It decomposes predictive changes into audio, video, and non-additive residual contributions, and adds only the selected contribution to the original audio-visual prediction.

\section{Method}%

\subsection{Preliminaries}

We consider autoregressive AV-LLMs, where each prediction is conditioned on paired audio-video inputs and a textual query.
Given audio $a$, video $v$, textual query $x$, and previously generated tokens $y_{<t}$, let $c_t=(x,y_{<t})$ denote the textual context at decoding step $t$.
For a candidate token $w\in\mathcal V$, where $\mathcal V$ is the vocabulary, an AV-LLM parameterized by $\theta$ assigns the next-token probability $p_\theta(w\mid a,v,c_t)$.
Throughout this work, the model parameters $\theta$ remain frozen, and all modifications are performed only at decoding time.

Contrastive decoding (CD) adjusts the original prediction using differences between predictions obtained under alternative input conditions~\citep{VCD_2024_CVPR}. For example, when the visual input is perturbed from $v$ to $v'$, a representative contrastive score $s_\mathrm{VCD}$ is
\begin{equation}
s_\mathrm{VCD}=(1+\alpha)\log p_\theta(w \mid a,v,c_t)
-
\alpha\log p_\theta(w \mid a,v',c_t),
\label{eq:cd}
\end{equation}
where $\alpha$ controls the contrastive strength. CD therefore specifies how predictions under two conditions are contrasted, while what the resulting difference represents depends on how the reference condition is constructed.

Among such difference-based formulations, pointwise mutual information (PMI)~\citep{Church_Hanks_1990_CL} is particularly useful when the goal is to quantify the predictive contribution of additional context. 
PMI measures how much observing \(z\) changes the probability assigned to \(w\) relative to its marginal probability (\ie, $\operatorname{PMI}(w;z)=\log p(w\mid z)-\log p(w)$).
Conditional PMI extends this comparison by holding a shared context fixed. For brevity, we sometimes refer to conditional PMI given $c_t$ simply as PMI when unambiguous.
In decoding, this corresponds to comparing next-token predictions with and without an additional source of context while keeping the base context fixed. 
Following prior works~\citep{CPMI_2025_NeurIPS,CAD_2024_NAACL}, we use the prediction obtained with the perceptual inputs omitted as a practical approximation to the corresponding marginal distribution.
Accordingly, treating the audio-visual input $(a,v)$ as the additional context, we obtain the following conditional PMI approximation:
\begin{equation}
\operatorname{PMI}(w; a,v \mid c_t)
\approx
\log p_\theta(w \mid a,v,c_t)
-
\log p_\theta(w \mid c_t).
\label{eq:pmi}
\end{equation}
The second term discounts predictive contribution already present in the question and generated context, so the difference captures the additional predictive contribution associated with the perceptual input. 
However, this joint formulation treats audio and video as a single source of additional context, leaving modality-specific and interaction contributions entangled.
\begin{figure*}[]
    \centering
    \includegraphics[width=0.95\textwidth]{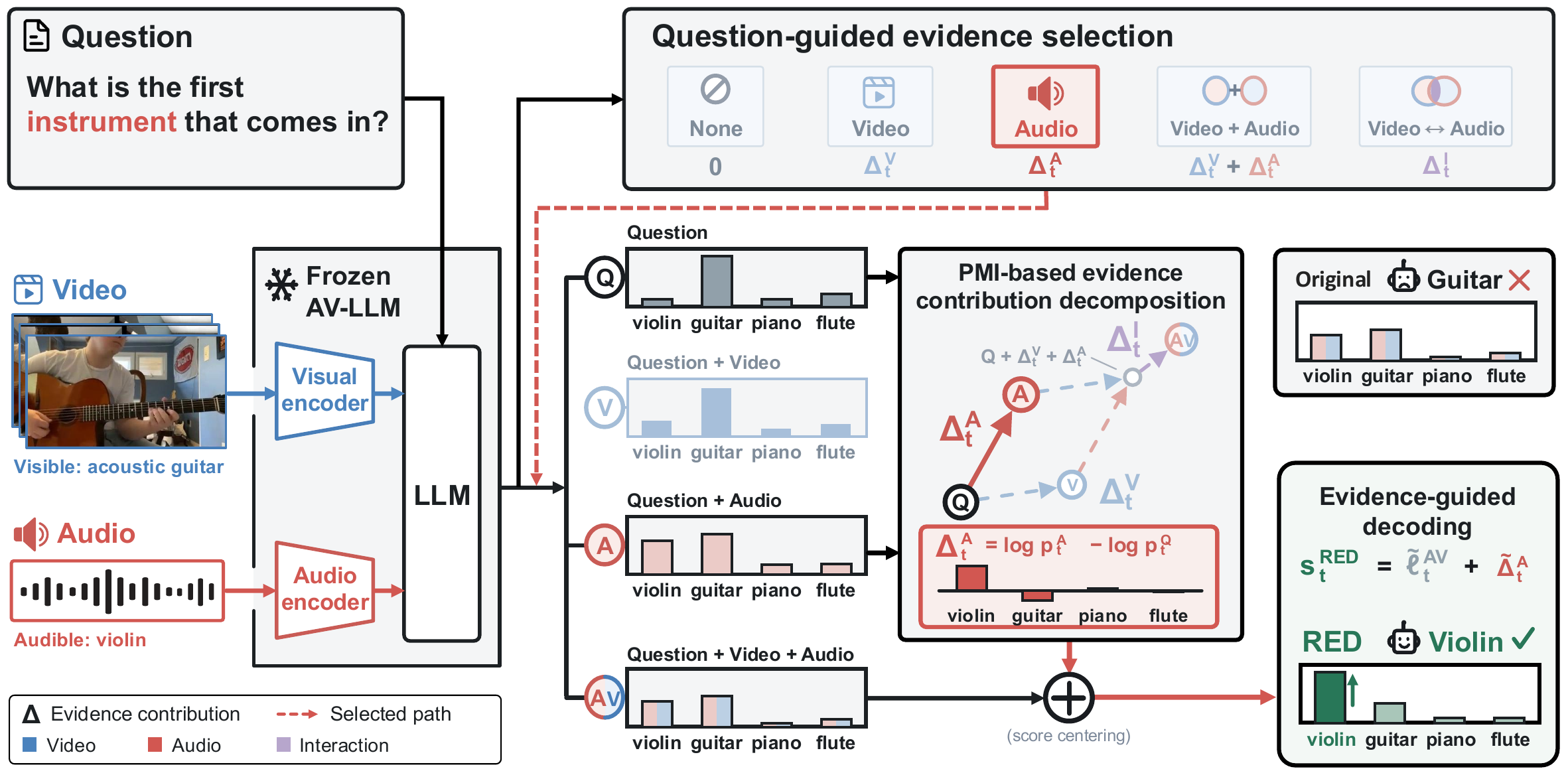}
    \caption{
    \textbf{Overview of \ours for an audio-required case.} 
(i) Given a query, \ours first identifies the evidence required to answer it, selecting among none, audio, video, audio + video, or audio-visual interaction.
(ii) The frozen AV-LLM produces the audio-visual prediction and only the auxiliary predictions required by the selected evidence type, from which RED measures audio, video, and non-additive audio-visual contributions relative to the textual reference.
(iii) The selected contribution augments the original audio-visual prediction to produce the final evidence-guided prediction.}
    \label{fig:method}
    \vspace{-1em}
\end{figure*}

\begin{wrapfigure}{r}{0.56\columnwidth}
\vspace{-1em}
    \centering

    \includegraphics[
        width=\linewidth,
        trim=2pt 2pt 2pt 2pt,
        clip
    ]{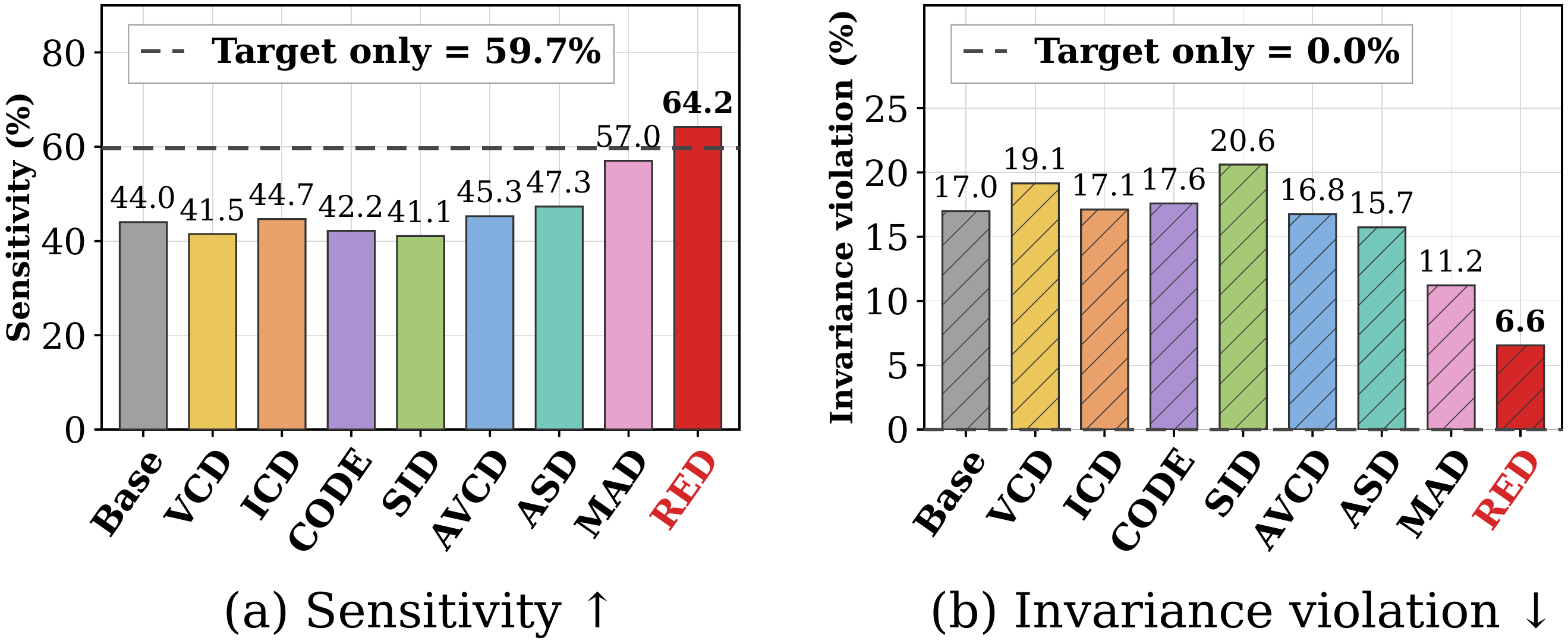}

    \vspace{-0.35em}

    \caption{
    \textbf{Evidence following under modality interventions.}
    Higher sensitivity and lower invariance violation indicate better evidence following.
    The dashed reference uses only the modality required by the question; its invariance violation is zero by construction because the intervened non-required modality is not used.
    Base receives both audio and video.
    Results are averaged over question-required evidence types and three backbones.}

    \label{fig:evidence_following}

    \vspace{-0.55em}
\end{wrapfigure}

\subsection{Motivation}

Entangled modality-specific and interaction effects raise a practical question:
\emph{does a contrastive correction actually make the prediction follow the evidence required by the question?}
Improved accuracy does not necessarily indicate stronger reliance on the required evidence; it may instead arise because non-required perceptual evidence moves the prediction toward the correct answer.

We therefore use controlled modality interventions to test whether predictions respond to required evidence while remaining stable to non-required evidence. 
We construct original–contrastive pairs from AVHBench by replacing one modality while keeping the question and the other modality fixed.
Replacing the required modality changes the correct answer and defines \emph{sensitivity}, whereas replacing a non-required modality preserves the answer and defines \emph{invariance violation}\footnote{Let $y,\hat{y}$ and $y',\hat{y}'$ denote the ground-truth answers and model predictions before and after intervention. Sensitivity counts $\mathbf{1}[\hat{y}=y \land \hat{y}'=y']$, while invariance violation counts $\mathbf{1}[\hat{y}\neq\hat{y}']$.}. Higher sensitivity and lower invariance violation indicate better evidence following.
Details of pair construction, replacement-sample verification, and metric computation are provided in Appendix~\ref{app:intervention_details}.

We compare inference with both audio and video against inference using only the modality required by the question.
As shown in Figure~\ref{fig:evidence_following}, sensitivity drops from 59.7\% when only the required modality is provided to 44.0\% when both audio and video are provided, while non-required-modality interventions change 17.0\% of predictions.
Existing contrastive decoding methods show a similar pattern.
Although they can improve final prediction performance, their predictions may still respond weakly to changes in question-required evidence.
These results suggest that the required evidence is often available to the model but may not contribute sufficiently when multiple modalities are jointly available. 
These observations motivate a correction that directly targets the predictive influence of question-relevant evidence, rather than only the final prediction.   
To capture this influence, we use pointwise mutual information to measure how each evidence source changes the prediction relative to a common reference.
Building on these contributions, the question determines which contribution to selectively augment while retaining the original audio-visual prediction.

\subsection{Relevant Evidence Decoding}
\label{sec:red}

We introduce Relevant Evidence Decoding (\ours), a training-free method that selectively augments predictive contribution from question-required evidence.
Figure~\ref{fig:method} provides an overview of \ours.

\paragraph{PMI-based evidence contribution decomposition.}
A high modality-conditioned probability need not imply strong support from that modality: the candidate may already be favored by the question and language context~\citep{M3ID_2024_CVPR,ONLY_2025_ICCV}.
We therefore measure each contribution as a change from the same question-only prediction.

For a candidate token $w$, let
$p_t^S(w)=p_\theta(w\mid z_S,x,y_{<t})$
denote the next-token probability under input condition
$S\in\{AV,A,V,Q\}$,
where $z_{AV}=(a,v)$, $z_A=a$, $z_V=v$, and $z_Q=\varnothing$.
All conditions use the same question and generated prefix.
These conditions define the possible evidence contributions; which of them is evaluated for a given query is determined by the evidence selection described below.

We define the evidence contribution for
$S$ as
\begin{equation}
\Delta_t^S(w)
=
\log p_t^S(w)-\log p_t^Q(w).
\label{eq:evidence_contributions}
\end{equation}
Thus, $\Delta_t^A$ and $\Delta_t^V$ measure the prediction changes induced by audio and video separately, while $\Delta_t^{AV}$ measures their joint contribution.
We define the remaining difference as the residual audio-visual
contribution $\Delta_t^I$:
\begin{equation}
\begin{aligned}
\Delta_t^I(w)
&=
\Delta_t^{AV}(w)
-
\Delta_t^A(w)
-
\Delta_t^V(w)\\
&=
\log p_t^{AV}(w)
+
\log p_t^Q(w)
-
\log p_t^A(w)
-
\log p_t^V(w).\\
\end{aligned}
\label{eq:pmi_decomposition}
\end{equation}
Equivalently,
$\Delta_t^{AV}
=
\Delta_t^A+\Delta_t^V+\Delta_t^I$,
where $\Delta_t^I$ denotes the interaction contribution,
defined as the non-additive residual under joint conditioning
that is not explained by the separate audio and video contributions.

\paragrapht{Question-guided evidence selection.}
\ours performs a question-only forward pass to determine what perceptual evidence is required to answer the query.
We organize the evidence requirements hierarchically by the number of perceptual modalities needed: none, one, or two.
With no perceptual input required, the evidence type is $\varnothing$.
With one modality required, it is audio ($A$) or video ($V$).
With two modalities required, we distinguish $A+V$ from $A\leftrightarrow V$.
Using only the textual query $x$, the same frozen backbone selects one of five evidence types:
\begin{equation}
\mathcal{E}(x)
\in
\left\{
\varnothing,\,
A,\,
V,\,
A+V,\,
A\leftrightarrow V
\right\}.
\label{eq:evidence_selection}
\end{equation}

For questions requiring both modalities, $A+V$ denotes cases in which
information from both audio and video is necessary without explicitly
querying the relation between them. In contrast, $A\leftrightarrow V$
denotes cases that explicitly query an audio-video relation, such as
correspondence or temporal alignment. The selector is executed once per query before autoregressive decoding,
and the selected evidence type is kept fixed across all decoding steps.

The selected evidence type directly determines which predictive contribution is added to the audio-visual prediction.
Let $e=\mathcal{E}(x)$ denote the selected evidence type.
To avoid tuning hyperparameters on labeled data, we center the scores using
unlabeled samples, following prior label-free methods~\citep{PriDe_2024_ICLR,BatchCalibration_2024_ICLR}.
We denote the resulting centered audio-visual score and evidence contributions by $\widetilde{\ell}_t^{AV}(w)$ and $\widetilde{\Delta}_t^S(w)$, respectively.

\ours then selects the corresponding contribution as
\begin{equation}
\rho_t(w)=
\begin{cases}
0, & e=\varnothing,\\
\widetilde{\Delta}_t^e(w), & e\in\{A,V\},\\
\widetilde{\Delta}_t^A(w)+\widetilde{\Delta}_t^V(w), & e=A+V,\\
\widetilde{\Delta}_t^I(w), & e=A\leftrightarrow V.
\end{cases}
\label{eq:selected_contribution}
\end{equation}
The final \ours score is
\begin{equation}
s_t^{\mathrm{RED}}(w)
=
\widetilde{\ell}_t^{AV}(w)
+
\rho_t(w).
\label{eq:red_score}
\end{equation}
The resulting score replaces the standard audio-visual next-token score during
decoding, while all other decoding operations remain unchanged.
\ours retains the $AV$ branch as the decoding anchor and evaluates only
the auxiliary $Q$, $A$, and $V$ branches required by the selected evidence
type.

\section{Experiments}

\subsection{Experimental setup}
\label{sec:exp_setup}

\vspace{-2mm}
\paragraph{Backbones.}
Following prior works~\citep{MAD_2026_CVPR}, we employ Qwen2.5-Omni-7B~\citep{Qwen2_5_Omni_2025_arXiv} and VideoLLaMA2.1-7B-AV~\citep{VideoLLaMA2_2024_arXiv} as the main backbone models to enable fair comparisons with existing methods.
We additionally evaluate the proposed method with Qwen2.5-Omni-3B to further demonstrate its effectiveness across different model scales.
Note that we keep all model parameters frozen.

\vspace{-2mm}
\paragraph{Benchmarks.}
We evaluate \ours across three complementary audio-visual hallucination benchmarks.
We evaluate on CMM~\citep{CMM_2025_NeurIPS} and AVHBench~\citep{AVHBench_2025_ICLR}, both established in prior audio-visual hallucination studies~\citep{MAD_2026_CVPR}.
CMM examines modality dominance and spurious cross-modal correlations, whereas AVHBench evaluates hallucinations caused by inconsistent audio-visual evidence.
We extend this evaluation with SVHalluc~\citep{SVHalluc_2026_CVPR}, covering semantic and temporal hallucinations in speech-vision understanding.
For generalization, we additionally evaluate on MUSIC-AVQA~\citep{MUSIC_AVQA_2022_CVPR} and its challenging MUSIC-AVQA-Hard subset~\citep{Kim_2026_Interspeech}.
We provide details on each benchmark in Appendix~\ref{app:benchmark}.

\vspace{-2mm}
\paragraph{Baselines.}
We compare \ours with two groups of training-free decoding baselines:
(1) audio-visual hallucination methods, including AVCD~\citep{AVCD_2025_NeurIPS}, ASD~\citep{ASD_2026_ICML}, and MAD~\citep{MAD_2026_CVPR}; and
(2) vision-language hallucination methods adapted to audio-visual inputs, including VCD~\citep{VCD_2024_CVPR}, ICD~\citep{ICD_2024_FindingsACL}, CODE~\citep{CODE_2024_NeurIPS}, and SID~\citep{SID_2025_ICLR}.
\footnote{All methods are re-evaluated under the same deterministic protocol using the settings reported in the original papers or released implementations.}
Implementation details are provided in Appendix~\ref{app:baselines}.

\begin{table*}[t]
\caption{
\textbf{Performance on audio--visual hallucination benchmarks.}
Accuracy (\%) on CMM, AVHBench, and SVHalluc.
Vis., Aud., and Lang.\ denote the CMM visual-, audio-, and language-dominance subsets.
For SVHalluc, Sem.\ and Temp.\ average the semantic
(GSA, FGSA, CMSB) and temporal (TA, TF, CMTB) tasks.
Avg. is the unweighted mean over all tasks in each benchmark.
}
\label{tab:main_full}

\vspace{-.4em}
\centering
\fontsize{8.5pt}{9pt}\selectfont
\setlength{\tabcolsep}{0.85pt}
\renewcommand{\arraystretch}{1.04}

\begin{tabularx}{\textwidth}{
@{}
l
*{3}{>{\centering\arraybackslash}X}
>{\columncolor{avggray}\centering\arraybackslash}X
*{3}{>{\centering\arraybackslash}X}
>{\columncolor{avggray}\centering\arraybackslash}X
*{2}{>{\centering\arraybackslash}X}
>{\columncolor{avggray}\centering\arraybackslash}X
@{}
}

\toprule

\multirow[c]{2}{*}{\raisebox{-1.5ex}{\textbf{Method}}}
& \multicolumn{4}{c}{\textbf{CMM}}
& \multicolumn{4}{c}{\textbf{AVHBench}}
& \multicolumn{3}{c}{\textbf{SVHalluc}}
\\

\cmidrule(lr){2-5}
\cmidrule(lr){6-9}
\cmidrule(lr){10-12}

&
\mbox{Vis.} 
& \mbox{Aud.}
& \mbox{Lang.}
& \textbf{Avg.}
& ADV
& VDA
& AVM
& \textbf{Avg.}
& Sem.
& Temp.
& \textbf{Avg.}
\\

\midrule

\multicolumn{12}{c}{\textbf{VideoLLaMA2.1-7B-AV}} \\
\cmidrule(lr){1-12}

Base
& 73.5 & 84.0 & 72.5 & 82.8
& 80.4 & 76.5 & 57.8 & 71.6
& 63.0 & 43.5 & 53.3
\\

VCD~\citep{VCD_2024_CVPR}
& 70.8 & 84.5 & 73.8 & 81.5
& \textbf{80.7} & 74.6 & 56.3 & 70.6
& 64.6 & 43.6 & 54.1
\\

ICD~\citep{ICD_2024_FindingsACL}
& 73.5 & 84.2 & 72.5 & 82.6
& \textbf{80.7} & 77.0 & 57.8 & 71.8
& 63.2 & 43.4 & 53.3
\\

CODE~\citep{CODE_2024_NeurIPS}
& 72.5 & 84.0 & 72.0 & 82.4
& 80.5 & 76.0 & 57.8 & 71.4
& 63.2 & 43.1 & 53.2
\\

SID~\citep{SID_2025_ICLR}
& 65.8 & 83.5 & 73.0 & 79.0
& 80.4 & 70.4 & 55.6 & 68.8
& 62.7 & 43.0 & 52.8
\\

AVCD~\citep{AVCD_2025_NeurIPS}
& 73.5 & 82.5 & 70.8 & 81.5
& \textbf{80.7} & 76.3 & 56.9 & 71.3
& 63.5 & 43.4 & 53.5
\\

ASD~\citep{ASD_2026_ICML}
& 74.5 & 85.5 & 72.0 & 82.4
& 80.0 & 76.5 & 56.5 & 71.0
& 59.9 & 41.9 & 50.9
\\

MAD~\citep{MAD_2026_CVPR}
& 82.0 & 82.8 & \textbf{77.8} & 84.7
& 79.9 & 80.6 & 58.5 & 73.0
& 68.2 & 43.7 & 55.9
\\

\rowcolor{ourslavender}
\textbf{RED (Ours)}
& \textbf{83.5}
& \textbf{88.0}
& 76.2
& \cellcolor{redavggray}\textbf{85.6}
& 78.4
& \textbf{82.1}
& \textbf{65.2}
& \cellcolor{redavggray}\textbf{75.3}
& \textbf{69.3}
& \textbf{44.8}
& \cellcolor{redavggray}\textbf{57.1}
\\

\midrule

\multicolumn{12}{c}{\textbf{Qwen2.5-Omni-3B}} \\
\cmidrule(lr){1-12}

Base
& 69.2 & 84.2 & 76.5 & 83.3
& 80.4 & 73.3 & 62.9 & 72.2
& 73.8 & 39.5 & 56.7
\\

VCD~\citep{VCD_2024_CVPR}
& 73.2 & 82.0 & 76.2 & 83.3
& 78.6 & 72.0 & 69.7 & 73.4
& 72.8 & 41.5 & 57.2
\\

ICD~\citep{ICD_2024_FindingsACL}
& 68.5 & 84.0 & 75.8 & 82.6
& 80.4 & 73.5 & 63.6 & 72.5
& 73.9 & 40.1 & 57.0
\\

CODE~\citep{CODE_2024_NeurIPS}
& 69.5 & 85.2 & 77.2 & 83.5
& 80.7 & 70.6 & 66.5 & 72.6
& 72.1 & 40.6 & 56.3
\\

SID~\citep{SID_2025_ICLR}
& 71.2 & 78.0 & 77.0 & 82.4
& 80.1 & 67.8 & 67.4 & 71.8
& 68.1 & 43.3 & 55.7
\\

AVCD~\citep{AVCD_2025_NeurIPS}
& 70.0 & 85.0 & 75.5 & 82.9
& 79.8 & 73.5 & 65.6 & 72.9
& 73.6 & 41.4 & 57.5
\\

ASD~\citep{ASD_2026_ICML}
& 70.0 & 84.8 & 78.0 & 83.9
& 82.2 & 73.7 & \textbf{70.8} & 75.5
& 73.8 & 40.5 & 57.1
\\

MAD~\citep{MAD_2026_CVPR}
& 74.5 & \textbf{87.0} & 78.8 & 84.6
& 82.7 & 80.0 & 62.4 & 75.0
& 75.1 & 43.9 & 59.5
\\

\rowcolor{ourslavender}
\textbf{RED (Ours)}
& \textbf{75.8}
& \textbf{87.0}
& \textbf{81.5}
& \cellcolor{redavggray}\textbf{86.2}
& \textbf{83.6}
& \textbf{81.2}
& 70.7
& \cellcolor{redavggray}\textbf{78.5}
& \textbf{77.0}
& \textbf{44.1}
& \cellcolor{redavggray}\textbf{60.5}
\\

\midrule

\multicolumn{12}{c}{\textbf{Qwen2.5-Omni-7B}} \\
\cmidrule(lr){1-12}

Base
& 69.5 & 71.5 & 82.2 & 82.2
& 80.3 & 73.4 & 74.9 & 76.2
& 76.1 & 44.9 & 60.5
\\

VCD~\citep{VCD_2024_CVPR}
& 69.0 & 68.5 & 82.5 & 81.7
& 76.5 & 72.2 & 77.1 & 75.3
& 74.6 & 45.2 & 59.9
\\

ICD~\citep{ICD_2024_FindingsACL}
& 68.8 & 71.2 & 81.8 & 82.0
& 80.5 & 72.9 & 76.0 & 76.5
& 75.9 & 44.9 & 60.4
\\

CODE~\citep{CODE_2024_NeurIPS}
& 68.0 & 68.5 & 82.8 & 81.4
& 78.6 & 71.1 & 73.9 & 74.5
& 76.5 & 45.1 & 60.8
\\

SID~\citep{SID_2025_ICLR}
& 73.5 & 65.8 & \textbf{83.5} & 82.0
& 77.1 & 75.7 & \textbf{77.8} & 76.9
& 73.0 & \textbf{46.2} & 59.6
\\

AVCD~\citep{AVCD_2025_NeurIPS}
& 69.5 & 72.8 & 75.2 & 80.4
& 79.3 & 76.4 & 73.1 & 76.3
& 77.8 & 45.7 & 61.7
\\

ASD~\citep{ASD_2026_ICML}
& 70.0 & 74.2 & 82.0 & 82.7
& 84.0 & 78.9 & 70.1 & 77.7
& 75.3 & 44.9 & 60.1
\\

MAD~\citep{MAD_2026_CVPR}
& 77.0 & 83.2 & 83.0 & 86.1
& \textbf{84.3} & 78.8 & 76.4 & 79.8
& 81.4 & 45.1 & 63.2
\\

\rowcolor{ourslavender}
\textbf{RED (Ours)}
& \textbf{86.0}
& \textbf{88.8}
& 83.0
& \cellcolor{redavggray}\textbf{89.2}
& 84.0
& \textbf{85.5}
& 76.3
& \cellcolor{redavggray}\textbf{81.9}
& \textbf{81.9}
& 44.9
& \cellcolor{redavggray}\textbf{63.4}
\\

\bottomrule
\end{tabularx}

\vspace{-2mm}
\end{table*}

\subsection{Main results}
\label{sec:main_results}

\vspace{-2mm}
\paragraph{Results on CMM.}
Table~\ref{tab:main_full} shows that \ours consistently improves CMM across all three backbones, with the largest gain on Qwen2.5-Omni-7B from 82.2\% to 89.2\%.
Notably, the improvement is concentrated on the modality-dominance subsets: \ours achieves absolute gains of 16.5\% and 17.3\% on the Vis. and Aud. Dominance subsets, respectively, while changing performance on Lang. only marginally.
This pattern indicates that \ours is particularly effective when perceptual modalities compete to determine the prediction.

\vspace{-2mm}
\paragraph{Results on AVHBench.}
\ours exhibits a similar trend on AVHBench, outperforming the strongest competing decoding methods by 2.1--3.0\% across all three backbones.
The gain is especially consistent on VDA, where \ours improves over the corresponding base models by 5.6, 7.9, and 12.1\%.
Together with the CMM results, this concentration of gains suggests that \ours is most beneficial in settings where the contribution of perceptual evidence is easily disrupted by other modalities.

\vspace{-2mm}
\paragraph{Results on SVHalluc.}
\ours achieves the best SVHalluc accuracy across all three backbones, improving standard decoding from 53.3--60.5\% to 57.1--63.4\% and consistently outperforming competing decoding methods.
Unlike prior audio-visual hallucination benchmarks that primarily rely on environmental sounds as indicators of event occurrence, SVHalluc requires grounding semantically rich speech with corresponding visual evidence~\citep{SVHalluc_2026_CVPR}.
The consistent advantage of \ours in this more challenging speech--vision setting suggests that selectively targeting question-required predictive contributions remains effective beyond simpler acoustic-event grounding.

\begin{figure*}[t]
\centering

\begin{minipage}[t]{0.45\textwidth}
    \vspace{0pt}
    \centering
    \centering

\captionof{table}{
\textbf{General audio-visual QA with Qwen2.5-Omni-7B.}
Accuracy (\%) and relative TTFT ($\times$ Base) on MUSIC-AVQA (Full) and MUSIC-AVQA-Hard (Hard).
}
\label{tab:music_all_methods}

\vspace{-.4em}

\begingroup
\footnotesize
\setlength{\tabcolsep}{0.85pt}
\renewcommand{\arraystretch}{1.04}

\begin{tabularx}{\linewidth}{
@{}
l
*{4}{>{\centering\arraybackslash}X}
@{}
}
\toprule

\multirow[c]{2}{*}{\textbf{Method}}
& \multicolumn{2}{c}{\textbf{Full}}
& \multicolumn{2}{c}{\textbf{Hard}}
\\

\cmidrule(lr){2-3}
\cmidrule(lr){4-5}

&
\textbf{Acc.}
& \textbf{TTFT}
& \textbf{Acc.}
& \textbf{TTFT}
\\

\midrule

Base
& 64.5
& 1.0$\times$
& 31.9
& 1.0$\times$
\\

\midrule

VCD
& 64.5
& 2.8$\times$
& 32.4
& 2.8$\times$
\\

SID
& 62.9
& 1.4$\times$
& 31.5
& 1.4$\times$
\\

ICD
& 65.7
& 3.1$\times$
& 32.7
& 3.0$\times$
\\

CODE
& 63.7
& 2.7$\times$
& 32.3
& 2.7$\times$
\\

AVCD
& 64.6
& 2.0$\times$
& 31.7
& 2.0$\times$
\\

ASD
& 62.4
& 4.1$\times$
& 28.2
& 4.2$\times$
\\

MAD
& \textbf{67.1}
& 2.7$\times$
& 36.0
& 2.6$\times$
\\

\rowcolor{ourslavender}
\textbf{RED (Ours)}
& \textbf{67.1}
& 1.6$\times$
& \textbf{40.5}
& 1.7$\times$
\\

\bottomrule
\end{tabularx}
\endgroup

\end{minipage}
\hfill
\begin{minipage}[t]{0.50\textwidth}
     \vspace{0.55em}
    \centering

    \includegraphics[
        width=0.92\linewidth,
        trim=2pt 2pt 2pt 2pt,
        clip
    ]{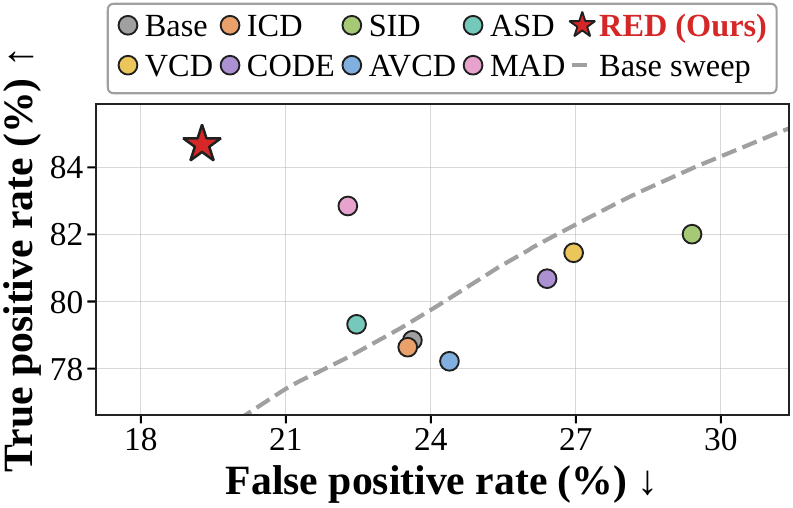}

    \vspace{-0.25em}

    \captionof{figure}{
    \textbf{True-positive and false-positive rates.}
    \ours achieves the highest TPR and the lowest FPR among the compared methods.
    The Base threshold sweep (dashed) does not attain \ours's TPR at the same or lower FPR.
    }
    \label{fig:tprfpr}
\end{minipage}

\vspace{-0.5em}
\end{figure*}

\vspace{-2mm}
\paragraph{Results on general audio-visual QA.}
Table~\ref{tab:music_all_methods} evaluates whether these gains extend beyond hallucination-specific benchmarks.
\ours improves Qwen2.5-Omni-7B from 64.5\% to 67.1\% on MUSIC-AVQA and from 31.9\% to 40.5\% on MUSIC-AVQA-Hard, matching MAD~\citep{MAD_2026_CVPR} on the former and exceeding it by 4.5\% on the latter.
The substantially larger gain on MUSIC-AVQA-Hard indicates that the benefit of \ours persists, and becomes more pronounced, as audio-visual reasoning becomes more challenging.

\subsection{Analyses}
\label{sec:analysis}

\vspace{-2mm}
\paragraph{True- and false-positive analysis.}
Accuracy on binary hallucination benchmarks can be confounded by response bias, making it difficult to determine whether performance gains arise from improved perceptual grounding or simply a shifted decision tendency~\citep{MIRAGEPerf_2025_NeurIPS}.
To disentangle these effects, we separately evaluate the true-positive rate (TPR) and false-positive rate (FPR), as shown in~\cref{fig:tprfpr}.
 
\ours improves both rates simultaneously, increasing the mean TPR from 78.8\% to 84.7\% while reducing the mean FPR from 23.6\% to 19.3\%.
Among methods that improve both rates relative to Base, \ours yields the largest improvement on each axis, reducing FPR by 4.3\% while increasing TPR by 5.9\%.
The dashed Base sweep varies the Base decision threshold on the same items, with the Base marker at threshold zero; no threshold reaches \ours's TPR at \ours's FPR.

\begin{wraptable}{r}{0.485\textwidth}
    \vspace{-0.8em}
    \centering

    \begin{minipage}{0.92\linewidth}
        \centering

\centering

\captionof{table}{
\textbf{Component analysis of evidence correction.}
We ablate question-guided evidence selection and PMI-based contribution
decomposition with the remaining RED procedure fixed.
Results are averaged over three backbones.
AUC denotes AUROC.
}
\label{tab:component_ablation}

\vspace{1em}

\begingroup
\footnotesize
\setlength{\tabcolsep}{0.85pt}
\renewcommand{\arraystretch}{1.04}

\begin{tabularx}{\linewidth}{
@{}
c
c
*{4}{>{\centering\arraybackslash}X}
@{}
}
\toprule

\multirow[c]{2}{*}{\textbf{Q-guided}}
&
\multirow[c]{2}{*}{\textbf{PMI}}
&
\multicolumn{2}{c}{\textbf{CMM}}
&
\multicolumn{2}{c}{\textbf{AVH}}
\\

\cmidrule(lr){3-4}
\cmidrule(lr){5-6}

&
&
\textbf{Acc.}
& \textbf{AUC}
& \textbf{Acc.}
& \textbf{AUC}
\\

\midrule

\xmark
& \xmark
& 84.8
& 92.3
& 75.2
& 82.0
\\

\cmark
& \xmark
& 86.8
& 93.6
& 76.2
& 82.7
\\

\xmark
& \cmark
& 84.2
& 91.9
& 75.8
& 82.8
\\

\rowcolor{ourslavender}
\cmark
& \cmark
& \textbf{87.0}
& \textbf{94.0}
& \textbf{78.6}
& \textbf{86.1}
\\

\bottomrule
\end{tabularx}
\endgroup

    \end{minipage}

    \vspace{-1.5em}
\end{wraptable}

\vspace{-2mm}

\paragraph{Component ablation.}
In \cref{tab:component_ablation}, we ablate question-guided evidence selection (\emph{Q-guided}) and PMI-based contribution decomposition (\emph{PMI}).
With only \emph{PMI} enabled, we add the audio, video, and residual interaction contributions to every question.
With only \emph{Q-guided} enabled, we replace PMI contributions with raw modality-conditioned scores: audio-only for Audio, video-only for Video, and the full audio-visual score for Both and Relation.
With neither component enabled, we add the raw audio-only and video-only scores to the audio-visual score.
Both components are necessary for the full \ours formulation: on AVHBench, adding PMI decomposition to Q-guided selection improves accuracy from 76.2\% to 78.6\% and AUROC from 82.7\% to 86.1\%.
The full model also achieves the strongest overall results, supporting the role of PMI decomposition in isolating the predictive contribution to be selectively reinforced.

\vspace{-2mm}
\paragraph{Evidence selection.}
To examine sensitivity to the selector implementation, Table~\ref{tab:route_ablation} compares Base, \emph{Selected target-only}, and alternative evidence assignments.
\emph{Selected target-only} uses the selected perceptual input, or full audio-visual input for Both and Relation, and substantially improves over Base; \ours retains the full audio-visual prediction and further improves it on both MUSIC-AVQA benchmarks.

For correction variants, only the evidence assignment changes:
\emph{Mismatched} averages the other three non-None types, \emph{Random} averages five uniform assignments, \emph{Heuristic} maps explicit question cues to evidence types, and \emph{Oracle} uses available question-level evidence labels as a label-informed upper-bound reference.
Mismatched and Random underperform \ours, whereas the competitive Heuristic and Oracle results show that correctly aligning the correction with question-required evidence is more important than the particular selector implementation.
\ours performs this selection with the same frozen backbone without an explicit rule-based mapping.

\begin{figure*}[!t]
    \centering

    \begin{minipage}[t]{0.475\textwidth}
        \vspace{0pt}
        \centering
        \centering

\captionof{table}{
\textbf{Evidence selection analysis on Qwen2.5-Omni-7B.}
Accuracy (\%) for Base, Selected target-only, and alternative evidence-assignment variants.
Full and Hard denote MUSIC-AVQA and MUSIC-AVQA-Hard.
}
\label{tab:route_ablation}

\begingroup
\fontsize{8pt}{9pt}\selectfont
\setlength{\tabcolsep}{0.85pt}
\renewcommand{\arraystretch}{1.04}

\begin{tabularx}{\linewidth}{
@{}
l
*{4}{>{\centering\arraybackslash}X}
@{}
}
\toprule

\multirow[c]{2}{*}{\textbf{Variant}}
&
\multirow[c]{2}{*}{\textbf{AVH}}
&
\multirow[c]{2}{*}{\textbf{CMM}}
&
\multicolumn{2}{c}{\textbf{MUSIC-AVQA}}
\\

\cmidrule(lr){4-5}

&
&
&
\textbf{Full}
&
\textbf{Hard}
\\

\midrule

Base
& 76.2
& 82.2
& 64.5
& 31.9
\\

Selected target-only
& 81.7
& 88.3
& 65.4
& 37.0
\\

\midrule

Mismatched
& 75.4
& 82.1
& 59.5
& 36.8
\\

Random
& 78.0
& 84.0
& 62.6
& 37.7
\\

Heuristic
& \textbf{81.9}
& \textbf{89.2}
& 66.9
& 40.7
\\

\rowcolor{ourslavender}
\textbf{RED}
& \textbf{81.9}
& \textbf{89.2}
& 67.1
& 40.5
\\

Oracle
& --
& --
& \textbf{67.7}
& \textbf{41.6}
\\

\bottomrule
\end{tabularx}
\endgroup

\vspace{0.15em}
\parbox{\linewidth}{
\scriptsize
``--'' denotes unavailable question-level oracle evidence labels.
}

        \label{tab:route_ablation}
    \end{minipage}
    \hfill
    \begin{minipage}[t]{0.475\textwidth}
        \vspace{0pt}
        \centering
        \includegraphics[
            width=0.92\linewidth,
            trim=2pt 2pt 2pt 0pt,
            clip
        ]{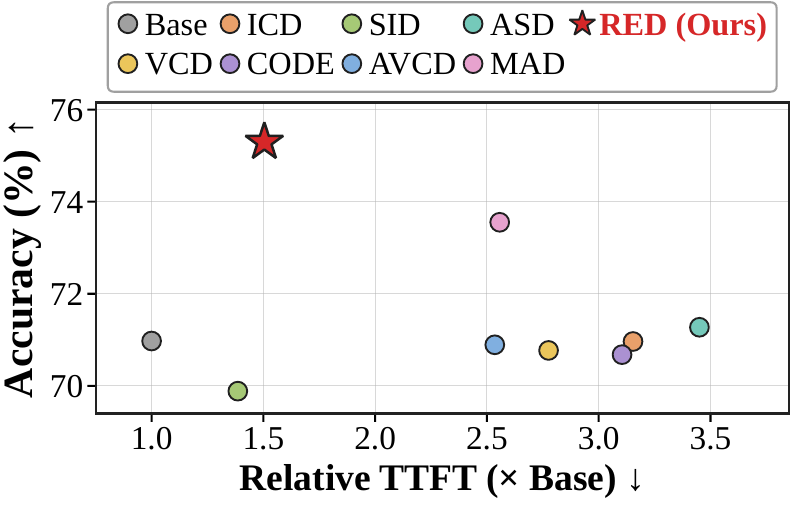}
\vspace{-1.5mm}
        \captionof{figure}{
        \textbf{Accuracy versus relative TTFT.} Means over CMM, AVHBench, and SVHalluc across three backbones. TTFT is normalized to the corresponding Base within each setting.
        }
        \label{fig:accuracy_cost}
    \end{minipage}

\end{figure*}

\begin{figure*}[t]
    \centering

    \begin{minipage}[t]{0.475\textwidth}
        \vspace{0pt}
        \centering

        \includegraphics[
            width=0.92\linewidth,
            trim=2pt 2pt 2pt 2pt,
            clip
        ]{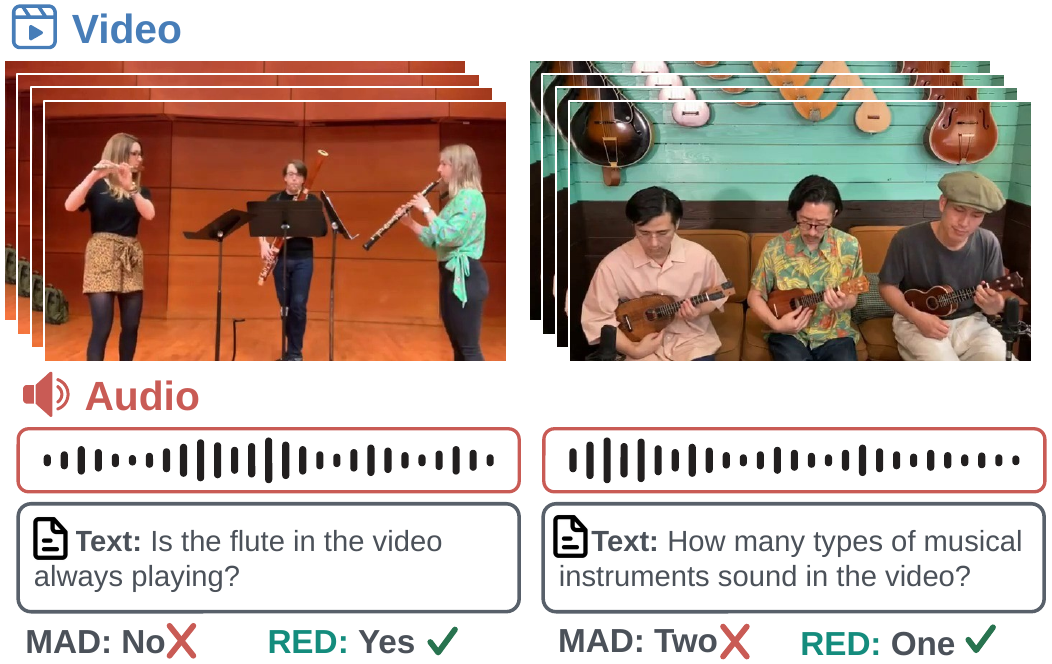}

        \vspace{-1.5mm}

        \captionof{figure}{
        \textbf{Qualitative examples on MUSIC-AVQA-Hard.}
        \ours correctly answers examples where MAD fails.
        }
        \label{fig:music_quality}
    \end{minipage}
    \hfill
    \begin{minipage}[t]{0.475\textwidth}
        \vspace{0pt}
        \centering

        \includegraphics[
            width=0.92\linewidth,
            trim=2pt 2pt 2pt 2pt,
            clip
        ]{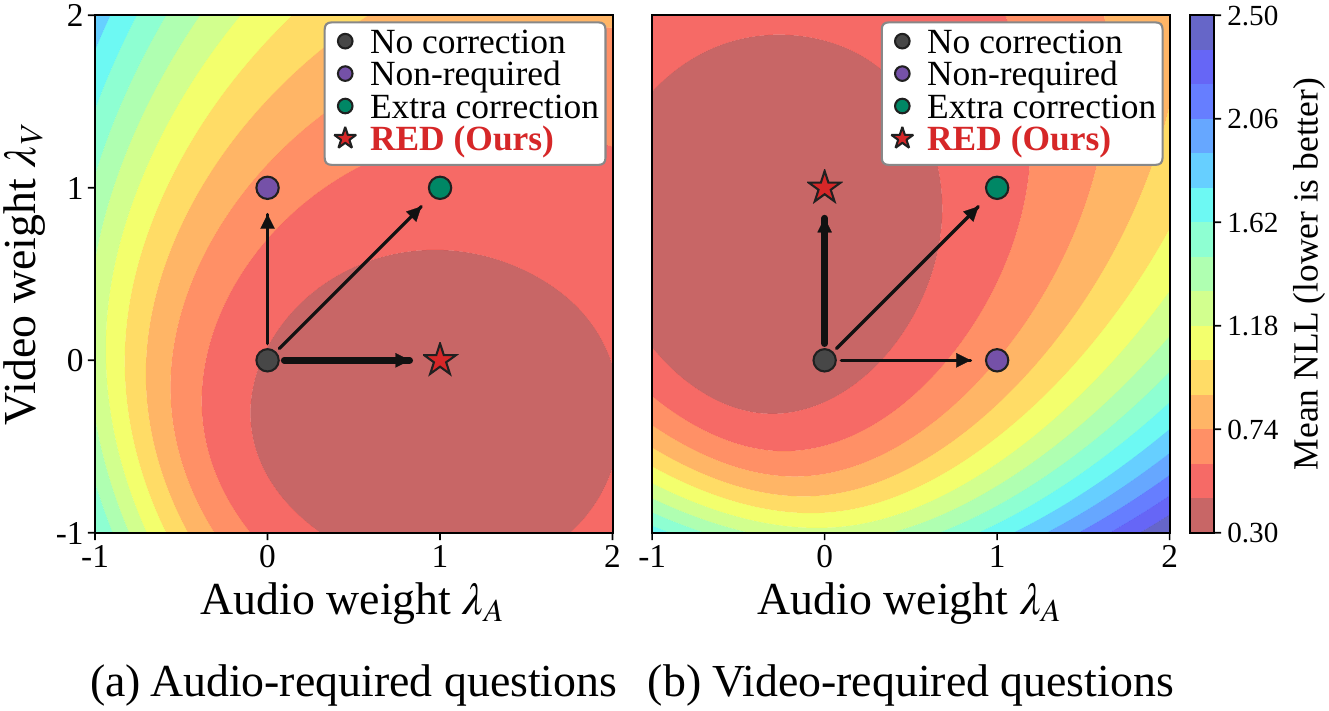}

        \vspace{-1.5mm}

        \captionof{figure}{
        \textbf{Loss landscape on CMM.} Mean binary-answer NLL over audio- and video-required subsets, averaged across three backbones. \ours corresponds to $(\lambda_A,\lambda_V)=(1,0)$ and $(0,1)$, respectively.}
        \label{fig:cmm_nll_landscape}
    \end{minipage}

    \vspace{-1.5em}
\end{figure*}
\vspace{-2mm}

\paragraph{Inference efficiency.}
Figure~\ref{fig:accuracy_cost} compares accuracy and relative TTFT averaged across nine backbone-benchmark settings.
\ours improves Base accuracy from 71.0\% to 75.3\% at 1.5$\times$ TTFT.
Compared with MAD~\citep{MAD_2026_CVPR}, \ours improves accuracy from 73.6\% to 75.3\% while reducing TTFT from 2.6$\times$ to 1.5$\times$, a 1.7$\times$ speedup.

\vspace{-2mm}

\paragraph{Qualitative and contribution analyses.}
Figure~\ref{fig:music_quality} illustrates modality bias in audio-visual decoding and shows that selectively augmenting question-required evidence can correct such failures without additional training.
Figure~\ref{fig:cmm_nll_landscape} further analyzes this behavior through the
loss landscape obtained by scaling the audio and video contributions with
$\lambda_A$ and $\lambda_V$, respectively, while keeping evidence selection fixed.
In both cases, \ours yields lower NLL than the three reference settings, supporting correction aligned with question-required evidence.
See Appendix~\ref{sec:evidence_contribution_analysis} for further details.

\section{Conclusion}
In this work, we study cross-modal hallucination in AV-LLMs through the lens of \emph{evidence following}, showing that correct predictions can remain weakly dependent on the perceptual evidence required by the question.
We introduce Relevant Evidence Decoding (\ours), a training-free method that decomposes audio-visual predictive changes relative to a common question-only reference and selectively augments the contribution corresponding to question-required evidence.
Across multiple AV-LLMs and benchmarks, \ours consistently improves hallucination mitigation and general audio-visual reasoning, while controlled interventions show increased sensitivity to required evidence and reduced sensitivity to irrelevant evidence.
These results suggest that effective multimodal decoding should not only improve prediction accuracy, but also ensure that predictions respond to the evidence that should determine the answer.

\subsection*{AI use statement}

During manuscript preparation, large language models were used for writing assistance to improve grammar, wording, and readability of text drafted by the authors. We also used feedback from Google’s Paper Assistant Tool (PAT) to support manuscript revision. The research ideas, methodological design, experimental setup, data analysis, and interpretation of results were carried out by the authors. The authors reviewed the final manuscript and take full responsibility for its content and claims.

\subsection*{Ethics statement}
Cross-modal hallucinations can reduce the reliability of multimodal systems,
particularly in applications such as assistive technologies, multimodal
assistants, and audio-visual content understanding.
By improving the use of question-relevant evidence, \ours may help make
existing AV-LLMs more dependable in real-world settings.
As a training-free method, \ours requires no task-specific retraining,
reducing the computational burden and limiting the need to use potentially
sensitive audio-visual data for additional model optimization.
These properties may facilitate hallucination mitigation in
resource-constrained and privacy-sensitive settings, supporting more reliable
use of existing multimodal systems.

\subsection*{Reproducibility statement}

Our experiments are conducted on NVIDIA RTX A6000 GPUs. Details required for reproducing the experiments, including dataset settings, evaluation protocols, baseline implementations, hyperparameter configurations, and the implementation details of \ours, are provided in the main text and Appendix. We plan to release the code and supporting implementation upon publication.

\bibliography{iclr2027_conference}
\bibliographystyle{iclr2027_conference}

\appendix
\clearpage
\appendix
\renewcommand\thefigure{\Alph{figure}}
\setcounter{figure}{0}
\renewcommand\thetable{\Alph{table}}
\setcounter{table}{0}

\noindent{\Large\textbf{Appendix}}\par\medskip
\suppressfloats[t]

\noindent
This appendix provides additional method details, experimental protocols, supplementary results, and qualitative analyses.
The contents are organized as follows:
\vspace{-0.3em}
\begin{itemize}
    \item Appendix~\ref{app:method_detail}: Method details and inference analysis;
    \item Appendix~\ref{app:implementation}: Experimental setup, baselines, benchmarks, and intervention details;
    \item Appendix~\ref{app:additional}: Additional quantitative results;
    \item Appendix~\ref{app:qualitative}: Qualitative analyses;
    \item Appendix~\ref{app:limitations}: Limitations.
\end{itemize}

\section{Method details}
\label{app:method_detail}

\subsection{Question-guided evidence selection}
\label{app:evidence_selection}

\begin{figure}[t]
\centering
\includegraphics[width=\linewidth]{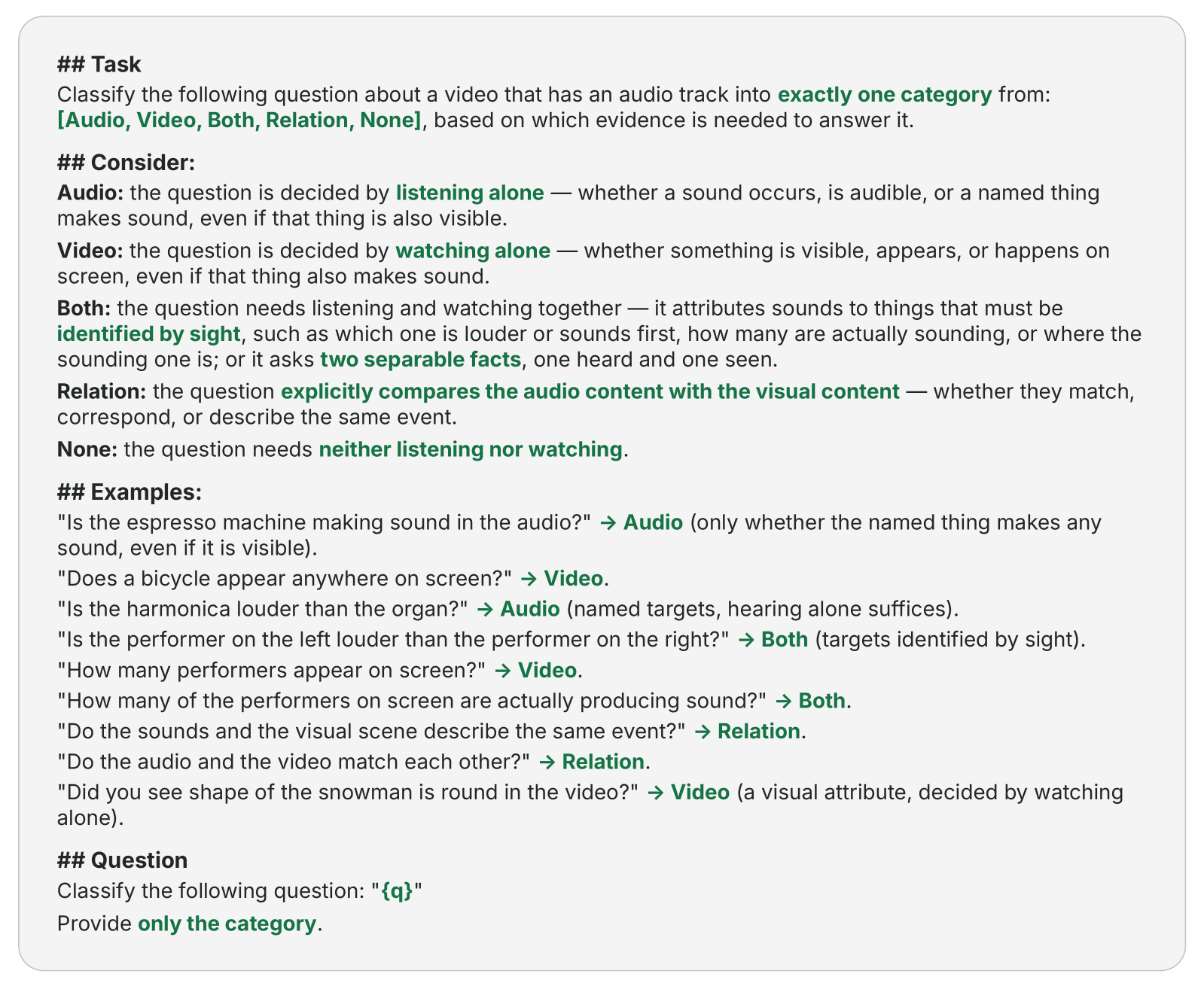}
\caption{\textbf{Question-guided evidence selection.} The frozen backbone identifies the required evidence using the question text alone.}
\label{fig:red_prompt}
\end{figure}

\ours determines the evidence required by the question using the same frozen backbone and the question text alone.
As shown in Figure~\ref{fig:red_prompt}, the model assigns each question to one of five evidence types---None, Audio, Video, Both, and Relation---corresponding to $\varnothing$, $A$, $V$, $A+V$, and $A\leftrightarrow V$ in the main text.
No audio or video input is provided during this selection step.
The selected evidence type determines which contribution is subsequently added to the audio-visual prediction.

\subsection{Effect of score centering}
\label{sec:centering_simple}

To match the sample budget used in prior decoding protocols, which select decoding strengths using 100 labeled examples per dataset~\citep{AVCD_2025_NeurIPS,MAD_2026_CVPR},
we use the same number of samples for score centering without labels.
Following prior label-free methods~\citep{PriDe_2024_ICLR,BatchCalibration_2024_ICLR},
we estimate fixed mean offsets from these unlabeled questions.
Let $\ell_t^{AV}(w)=\log p_t^{AV}(w)$ denote the uncentered audio-visual log-probability score.
The centered scores are
$\widetilde{\ell}_t^{AV}(w)=\ell_t^{AV}(w)-\eta_e^{AV}(w)$ and
$\widetilde{\Delta}_t^S(w)=\Delta_t^S(w)-\eta_e^S(w)$,
where $S\in\{A,V,I\}$.
Here, $e$ denotes the selected evidence type, and $\eta_e^{AV}(w)$ and $\eta_e^S(w)$ are the means of the corresponding audio-visual score and evidence contribution for candidate $w$, computed over the sampled questions assigned to $e$ for each backbone and benchmark.
When fewer than five valid samples are available, we use the mean of the corresponding quantity pooled across evidence types.
The offsets are kept fixed throughout evaluation.

Table~\ref{tab:centering_simple} isolates the effect of mean-offset centering on CMM while fixing the evidence assignments, contribution definitions, and evaluation population.
Centering improves accuracy from 87.3\% to 89.2\% on Qwen2.5-Omni-7B, while producing smaller changes on Qwen2.5-Omni-3B and VideoLLaMA2.1-7B-AV.
In contrast, AUROC changes only marginally across all three backbones, from 95.4\% to 95.5\%, 93.6\% to 93.6\%, and 92.5\% to 92.8\%, respectively.
This pattern suggests that centering primarily corrects backbone-dependent systematic score offsets at the final decision stage, rather than substantially changing the relative discrimination induced by the evidence-based scores.

\begin{table}[t]
\caption{\textbf{Effect of score centering on CMM.} Task-macro accuracy and AUROC (\%) for \ours with and without centering.}
\label{tab:centering_simple}
\vspace{-.4em}
\centering
\footnotesize
\setlength{\tabcolsep}{0.85pt}
\renewcommand{\arraystretch}{1.04}
\begin{tabularx}{0.50\linewidth}{@{}l*{3}{>{\centering\arraybackslash}X}@{}}
\toprule
\textbf{Method} & \textbf{Centering} & \textbf{Acc.} & \textbf{AUROC} \\
\midrule
\multicolumn{4}{c}{\textbf{VideoLLaMA2.1-7B-AV}} \\
\cmidrule(lr){1-4}
RED (Ours) & $\times$ & 85.5 & 92.5 \\
\rowcolor{ourslavender}
\textbf{RED (Ours)} & $\checkmark$ & \textbf{85.6} & \textbf{92.8} \\
\midrule
\multicolumn{4}{c}{\textbf{Qwen2.5-Omni-3B}} \\
\cmidrule(lr){1-4}
RED (Ours) & $\times$ & 86.1 & \textbf{93.6} \\
\rowcolor{ourslavender}
\textbf{RED (Ours)} & $\checkmark$ & \textbf{86.2} & \textbf{93.6} \\
\midrule
\multicolumn{4}{c}{\textbf{Qwen2.5-Omni-7B}} \\
\cmidrule(lr){1-4}
RED (Ours) & $\times$ & 87.3 & 95.4 \\
\rowcolor{ourslavender}
\textbf{RED (Ours)} & $\checkmark$ & \textbf{89.2} & \textbf{95.5} \\
\bottomrule
\end{tabularx}
\end{table}

\begin{figure*}[t]
      \centering
      \includegraphics[width=\textwidth]
      {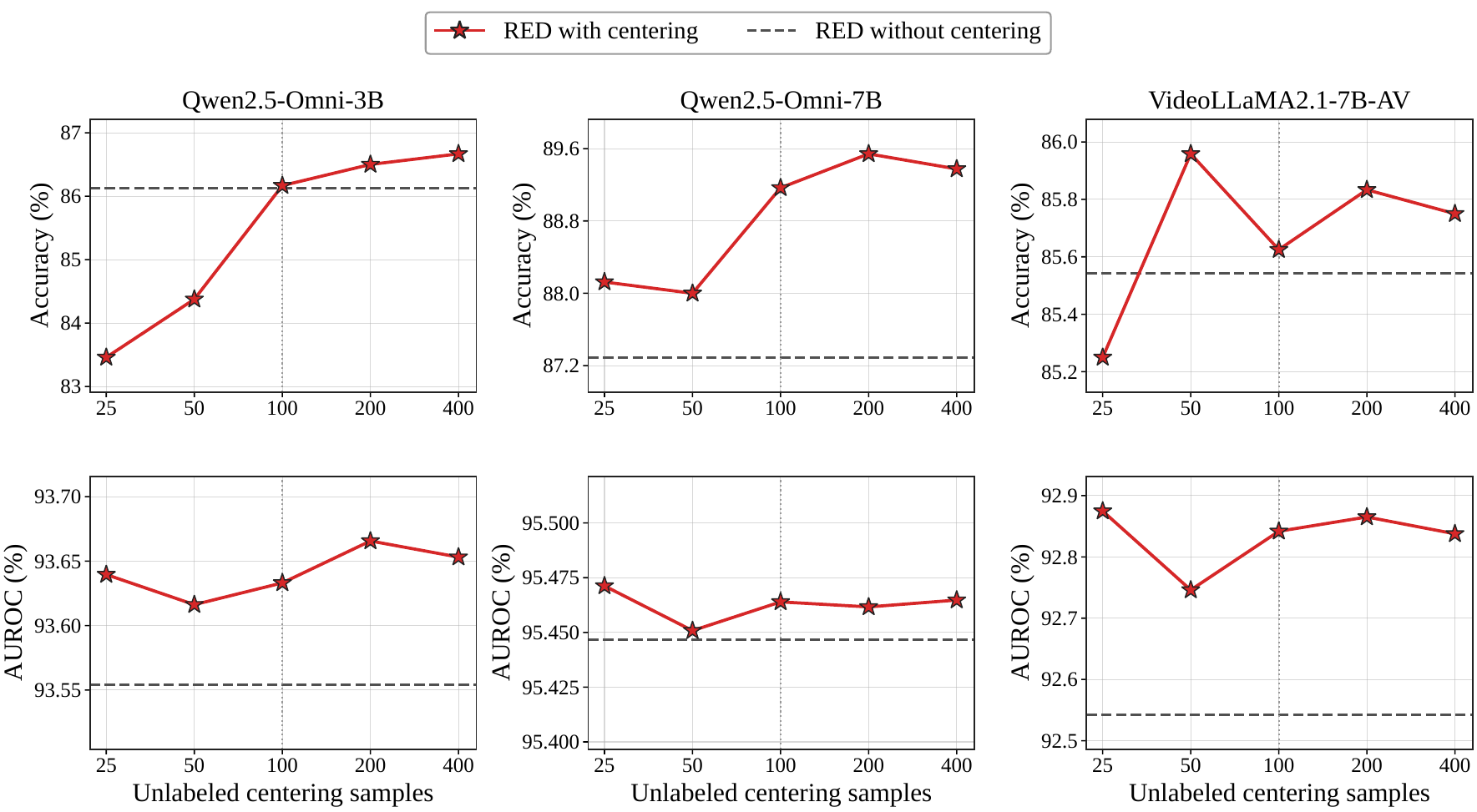}
      \caption{
      \textbf{Sensitivity to centering sample size on CMM.}
      We vary the number of unlabeled centering questions under a fixed
      sampling configuration.
      Dashed lines denote \ours without centering; the dotted vertical line
      marks the default budget of 100 questions.
      }
      \label{fig:centering_budget}
  \end{figure*}

\paragraph{Sensitivity to sample size.}
  Figure~\ref{fig:centering_budget} examines the effect of using
  25, 50, 100, 200, or 400 unlabeled questions to estimate the centering offsets
  on CMM across three backbones.
  We keep the sampling seed, evidence assignments, and backbone predictions
  fixed, varying only the sample budget.
  Increasing the budget does not improve performance monotonically:
  Qwen2.5-Omni-3B benefits from larger budgets, while Qwen2.5-Omni-7B
  and VideoLLaMA2.1-7B-AV show smaller accuracy changes beyond 100 questions.

\begin{figure*}[t]
      \centering
      \includegraphics[width=\textwidth]
      {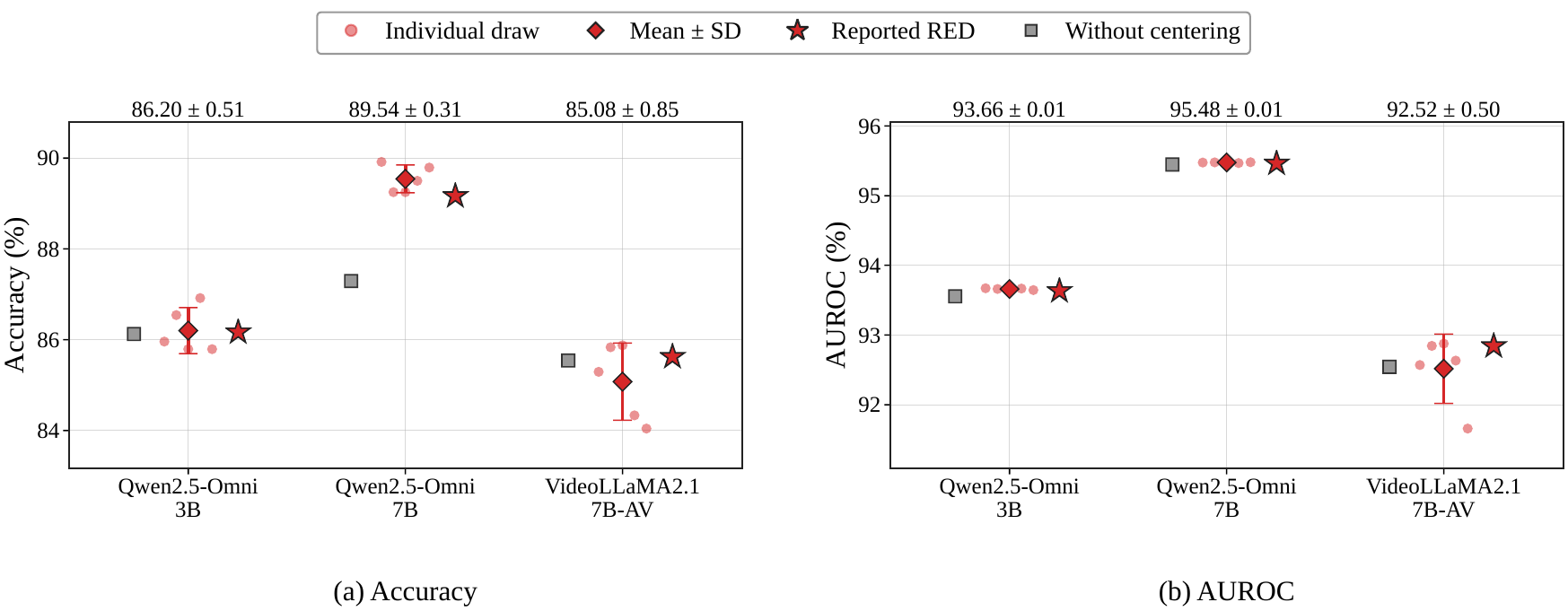}
      \caption{
      \textbf{Robustness to centering sample selection on CMM.}
      Small circles show results from five random draws of 100 unlabeled
      questions.
      Diamonds and error bars indicate the mean and one sample standard
      deviation.
      Stars show the reported \ours results, and squares denote \ours
      without centering.
      }
      \label{fig:centering_resampling}
  \end{figure*}

  \paragraph{Robustness to sample selection.}
  To assess variability at the default budget, we independently sample
  100 unlabeled questions using five random seeds and recompute the offsets
  for each draw, keeping all other settings fixed.
  Following our evaluation protocol, each draw is evaluated on all
  2,400 CMM questions, including the sampled questions.
  Figure~\ref{fig:centering_resampling} reports the individual results and
  their mean and standard deviation alongside the reported \ours results.
  Accuracy is $86.20\pm0.51\%$, $89.54\pm0.31\%$, and $85.08\pm0.85\%$
  for Qwen2.5-Omni-3B, Qwen2.5-Omni-7B, and VideoLLaMA2.1-7B-AV,
  respectively.
  The accuracy improvement from centering persists across all five draws
  for Qwen2.5-Omni-7B, whereas the other backbones show less consistent
  benefits, with greater sampling variability for VideoLLaMA2.1-7B-AV.

\subsection{Diagnostics of evidence contribution selection}
\label{sec:evidence_contribution_analysis}

\ours associates each selected evidence type with a corresponding centered predictive contribution: $A\!\rightarrow\!\widetilde{\Delta}^A$, $V\!\rightarrow\!\widetilde{\Delta}^V$, $A+V\!\rightarrow\!\widetilde{\Delta}^A+\widetilde{\Delta}^V$, and $A\leftrightarrow V\!\rightarrow\!\widetilde{\Delta}^I$.
For brevity, we omit the tildes in the figures and tables of this subsection; all reported correction terms use the centered contributions defined in the main text.
We examine whether prediction quality depends on which contribution is added to the audio-visual score.
Within each comparison, we keep the question-guided evidence selection, auxiliary predictions, and centering offsets fixed, and vary only the added contribution.
This isolates the effect of contribution choice from changes in evidence selection.

\begin{figure*}[t]
\centering
\includegraphics[width=\textwidth]{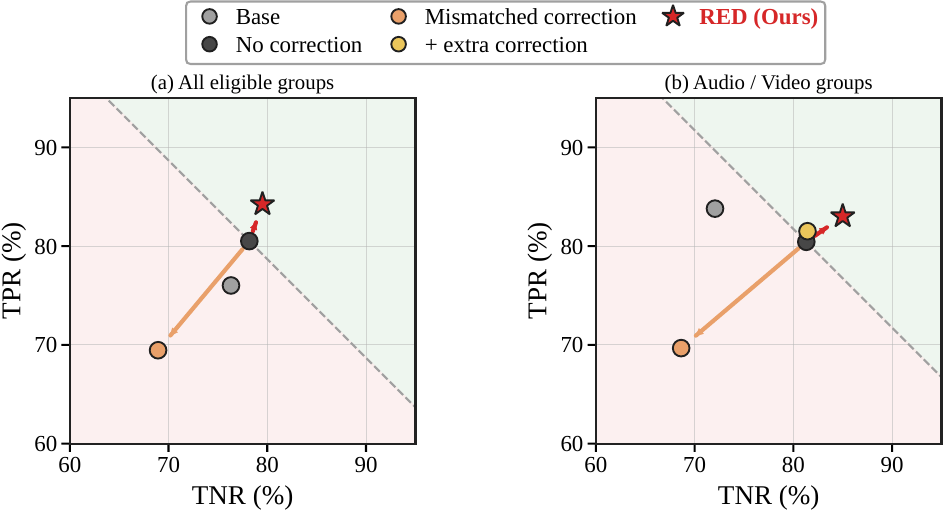}
\caption{
\textbf{Effect of choosing different evidence contributions.}
We keep the evidence selection and centering fixed and vary only the contribution added to the audio-visual score.
The left panel includes all eligible groups, while the right panel focuses on Audio and Video groups and additionally compares adding the other unimodal contribution.
\emph{No correction} denotes the centered audio-visual score without an additional evidence contribution.
Higher TPR and TNR are better.
}
\label{fig:contribution_choice}
\end{figure*}

\paragraph{Contribution choice matters.}
Figure~\ref{fig:contribution_choice} examines how the choice of evidence contribution affects the true-positive and true-negative operating point.
For all correction-based variants, we keep the question-guided evidence selection, auxiliary predictions, and centering procedure fixed and vary only the contribution added to the audio-visual score.
\emph{Base} denotes standard decoding without score centering or an additional evidence contribution.
\emph{No correction} uses the centered audio-visual score without an additional evidence contribution, whereas \ours adds the contribution corresponding to the selected evidence type.
\emph{Mismatched correction} keeps the selected evidence type fixed but adds a non-corresponding contribution; this differs from the mismatched evidence-assignment strategy analyzed in the main text.

Across the eligible groups, \ours moves the operating point toward higher TPR and TNR, whereas mismatched corrections move it in the opposite direction.
For Audio and Video groups, we additionally consider \emph{+ extra correction}, which adds both unimodal contributions rather than only the selected one.
Its limited benefit compared with \ours indicates that adding more modality contributions is not necessarily beneficial; the correction should match the evidence required by the question.

\paragraph{Loss landscape on CMM.}
Figure~\ref{fig:cmm_nll_landscape} shows binary-answer negative log-likelihood (NLL)
as the audio and video contributions are scaled.
For each question, we form
\[
s_{\lambda_A,\lambda_V}(w)
=
\widetilde{\ell}^{AV}(w)
+\lambda_A\widetilde{\Delta}^{A}(w)
+\lambda_V\widetilde{\Delta}^{V}(w),
\qquad w\in\{\mathrm{Yes},\mathrm{No}\},
\]
and compute the NLL of the reference answer after normalizing these
two scores with a softmax.
We vary $\lambda_A$ and $\lambda_V$ over $[-1,2]$ with a step of 0.025,
while keeping the evidence selections, underlying predictions,
and centering offsets fixed.
The offsets are estimated from 100 unlabeled CMM questions using
sampling seed 5, following Appendix~\ref{sec:centering_simple}.

The analysis uses questions with all four prediction branches
available from the VAL, Vis., and Aud.\ subsets of CMM.
All four branches are required because the two-dimensional sweep evaluates both \(\widetilde{\Delta}^{A}\) and \(\widetilde{\Delta}^{V}\) for the same question, including the
non-selected contribution that \ours would not normally evaluate.
We group these questions by each backbone's question-only evidence
selection, retaining the Audio and Video groups.
For Qwen2.5-Omni-3B, Qwen2.5-Omni-7B, and VideoLLaMA2.1-7B-AV,
respectively, the Audio groups contain 618, 600, and 513 questions,
and the Video groups contain 582, 600, and 686 questions.
At each coefficient pair, we first average NLL over questions
within each backbone's group and then average equally across
the three backbones.

The no-correction setting is $(\lambda_A,\lambda_V)=(0,0)$
and uses only the centered audio-visual score.
\ours corresponds to $(1,0)$ for Audio and $(0,1)$ for Video;
non-required correction uses the opposite modality contribution,
and extra correction corresponds to $(1,1)$.
Note that \ours uses the fixed coefficients implied by its decoding rule;
the landscape is a diagnostic analysis and is not used to tune or select
decoding strengths.

\paragraph{Modality-specific and residual contributions.}
Table~\ref{tab:avh_three_deltas} examines Audio, Video, and Relation evidence requirements on AVHBench.
For VDA, which requires audio evidence, $\Delta^A$ achieves the highest accuracy and AUROC.
For ADV, which requires video evidence, $\Delta^V$ performs best on both metrics.
For AVM, which evaluates the relation between audio and visual content, the residual audio-visual contribution $\Delta^I$ achieves the highest accuracy and AUROC.
The preferred contribution therefore matches the corresponding $A$, $V$, and $A\leftrightarrow V$ evidence requirements.

\begin{table*}[t]
\caption{
\textbf{Modality-specific and residual contributions on AVHBench.}
Accuracy and AUROC (\%) with Qwen2.5-Omni-3B when adding the audio ($\Delta^A$), video ($\Delta^V$), or residual audio-visual ($\Delta^I$) contribution to the same centered audio-visual score.
VDA, ADV, and AVM correspond to Audio ($A$), Video ($V$), and Relation ($A\leftrightarrow V$) evidence requirements, respectively.
The centering offsets are fixed across variants.
\textbf{Bold} indicates the best result in each row for each metric.
}
\label{tab:avh_three_deltas}
\vspace{-.4em}
\centering
\footnotesize
\setlength{\tabcolsep}{0.85pt}
\renewcommand{\arraystretch}{1.04}
\begin{tabularx}{\textwidth}{@{}ll*{6}{>{\centering\arraybackslash}X}@{}}
\toprule
\multirow[c]{2}{*}{\raisebox{-1.5ex}{\textbf{Task}}} &
\multirow[c]{2}{*}{\raisebox{-1.5ex}{\textbf{Evidence}}} &
\multicolumn{2}{c}{$+\Delta^A$} & \multicolumn{2}{c}{$+\Delta^V$} & \multicolumn{2}{c}{$+\Delta^I$} \\
\cmidrule(lr){3-4}\cmidrule(lr){5-6}\cmidrule(lr){7-8}
& & \textbf{Acc.} & \textbf{AUROC} & \textbf{Acc.} & \textbf{AUROC} & \textbf{Acc.} & \textbf{AUROC} \\
\midrule
VDA & $A$ & \textbf{81.27} & \textbf{89.24} & 70.39 & 76.89 & 46.38 & 44.60 \\
ADV & $V$ & 77.99 & 84.26 & \textbf{83.63} & \textbf{88.82} & 64.08 & 68.20 \\
AVM & $A\leftrightarrow V$ & 69.56 & 77.65 & 62.42 & 68.51 & \textbf{70.68} & \textbf{78.54} \\
\bottomrule
\end{tabularx}
\end{table*}

\paragraph{Additive contributions for both-modality evidence.}
The $A+V$ evidence type is too sparsely represented on AVHBench for a meaningful direct analysis.
We therefore examine this case on MUSIC-AVQA-Hard, which also provides dataset evidence annotations identifying questions that require both audio and video.

Table~\ref{tab:music_both_deltas} compares questions selected as $A+V$ by \ours's question-guided evidence selection with questions identified by the dataset's Audio-Visual evidence tags.
On the Oracle-tagged subset, $\Delta^A+\Delta^V$ achieves the highest accuracy at 40.16\%, outperforming either modality-specific contribution alone.
This result supports the additive contribution when the evidence requirement explicitly involves both modalities.

The Prompt-selected subset exhibits a different pattern: $\Delta^V$ slightly outperforms $\Delta^A+\Delta^V$ (38.63\% vs.\ 38.19\%).
Because \ours determines the required evidence from the textual query alone, the questions it assigns to $A+V$ need not coincide with those identified by the dataset annotations.
Accordingly, the effectiveness of selective decoding depends not only on mapping an evidence type to its predictive contribution, but also on identifying the evidence required by the question.

\begin{table*}[t]
\caption{
\textbf{Additive contributions for both-modality evidence on MUSIC-AVQA-Hard.}
Accuracy (\%) with Qwen2.5-Omni-7B.
Prompt-selected contains questions assigned to $A+V$ by \ours's question-guided evidence selection, whereas Oracle-tagged contains questions identified by the dataset's Audio-Visual evidence tags.
All variants use the same centering offsets and differ only in the contribution added to the centered audio-visual score.
\textbf{Bold} indicates the best result in each row.
}
\label{tab:music_both_deltas}
\vspace{-.4em}
\centering
\footnotesize
\setlength{\tabcolsep}{0.85pt}
\renewcommand{\arraystretch}{1.04}
\begin{tabularx}{\textwidth}{@{}l*{5}{>{\centering\arraybackslash}X}@{}}
\toprule
\textbf{$A+V$ subset} & $\mathbf{n}$ & $+\Delta^A$ & $+\Delta^V$ & $+(\Delta^A+\Delta^V)$ & $+\Delta^I$ \\
\midrule
Prompt-selected & 1,786 & 36.73 & \textbf{38.63} & 38.19 & 33.82 \\
Oracle-tagged & 2,124 & 37.95 & 38.79 & \textbf{40.16} & 30.23 \\
\bottomrule
\end{tabularx}
\end{table*}

\begin{figure*}[t]
\centering
\includegraphics[width=0.95\textwidth]{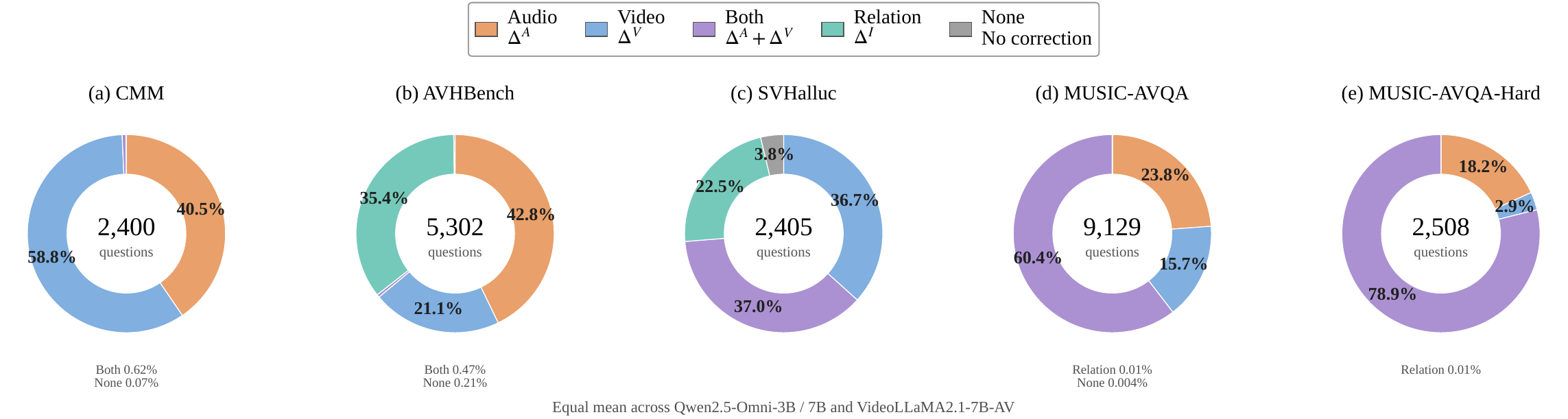}
\caption{
\textbf{Question-guided evidence selection across five benchmarks.}
For each backbone, proportions are computed over all questions in each evaluation split and then averaged equally across three backbones.
Centers report the number of questions per backbone.
Audio, Video, Both, Relation, and None select $\Delta^A$, $\Delta^V$, $\Delta^A+\Delta^V$, $\Delta^I$, and no additional correction, respectively.
Both selects additive audio and video contributions; Relation selects their residual interaction.
The distributions reflect question-only selector predictions, not oracle annotations.
Nonzero categories below 1\% are listed beneath each chart.
MUSIC-AVQA uses the official corrected test annotations (9,129 questions); MUSIC-AVQA-Hard contains 2,508 questions from this split.
}
\label{fig:contribution_selection_distribution}
\end{figure*}

\paragraph{Evidence selection across benchmarks.}
Figure~\ref{fig:contribution_selection_distribution} summarizes question-only evidence selections across five benchmarks, averaged equally over Qwen2.5-Omni-3B, Qwen2.5-Omni-7B, and VideoLLaMA2.1-7B-AV.
Each backbone's distribution includes all questions in the corresponding evaluation split: 2,400 for CMM, 5,302 for AVHBench, 2,405 for SVHalluc, 9,129 for the officially corrected MUSIC-AVQA test split, and 2,508 for MUSIC-AVQA-Hard.
CMM predominantly selects Audio or Video, while Relation accounts for a substantial share on AVHBench.
SVHalluc shows a broader distribution over Video, Both, and Relation.
Both is the most frequently selected type on MUSIC-AVQA and MUSIC-AVQA-Hard, especially on the latter.
These distributions describe the selectors' predicted evidence requirements rather than ground-truth modality dependencies.

\subsection{Inference Efficiency Analysis}
\label{app:inference_efficiency}

\paragraph{Inference stages.}
Table~\ref{tab:red_runtime_stages} profiles \ours across three backbones on CMM, AVHBench, and SVHalluc.
The selector predicts the evidence type once per question, while the required answer-prediction paths are evaluated separately.
Evidence selection incurs little overhead, whereas media
preparation and perceptual encoding account for most of the runtime.
By evaluating only the auxiliary prediction paths required by the selected evidence type, \ours avoids unnecessary perceptual processing.

\begin{table}[t]
\centering
\caption{
  \textbf{Runtime breakdown of \ours.}
  Stage times are averaged equally over CMM, AVHBench, and
  SVHalluc across three backbones using separate profiling runs.
  }

\label{tab:red_runtime_stages}
\begingroup
  \footnotesize
  \setlength{\tabcolsep}{3.2pt}
  \renewcommand{\arraystretch}{1.05}
  \begin{tabular}{@{}lr@{}}
  \toprule
  \textbf{Stage} & \textbf{Time (ms)} \\
  \midrule
  Evidence selection & 89.1 \\
  Media preparation & 2372.1 \\
  Perceptual encoding & 1881.4 \\
  Answer-prompt prefill & 1381.7 \\
  Answer-token prediction & 28.4 \\
  Score composition and overhead & 37.2 \\
  \midrule
  \rowcolor{ourslavender}
  \textbf{RED (Ours)} & \textbf{5789.9} \\
  \bottomrule
  \end{tabular}
  \endgroup
  \end{table}

\paragraph{TTFT across benchmarks.}
  Table~\ref{tab:latency_detailed} reports relative TTFT for each
  backbone and hallucination benchmark.
  \ours requires approximately $1.2\times$--$1.7\times$ Base TTFT
  and remains faster than MAD and ASD across all nine settings,
  consistent with evaluating only the auxiliary prediction paths
  required by the selected evidence type.

\begin{table*}[t]
\caption{
  \textbf{Relative TTFT across audio-visual hallucination benchmarks.}
  Time to first token normalized to the corresponding Base
  ($\times$ Base; lower is better).
}
\label{tab:latency_detailed}
\vspace{-.4em}
\centering
\footnotesize
\setlength{\tabcolsep}{0.85pt}
\renewcommand{\arraystretch}{1.04}

\begin{tabularx}{\textwidth}{
@{}
l
*{9}{>{\centering\arraybackslash}X}
@{}
}

\toprule

\multirow[c]{2}{*}{\raisebox{-1.5ex}{\textbf{Method}}}
& \multicolumn{3}{c}{\textbf{VideoLLaMA2.1-7B-AV}}
& \multicolumn{3}{c}{\textbf{Qwen2.5-Omni-3B}}
& \multicolumn{3}{c}{\textbf{Qwen2.5-Omni-7B}}
\\

\cmidrule(lr){2-4}
\cmidrule(lr){5-7}
\cmidrule(lr){8-10}

& \textbf{CMM}
& \textbf{AVH}
& \textbf{SVH}
& \textbf{CMM}
& \textbf{AVH}
& \textbf{SVH}
& \textbf{CMM}
& \textbf{AVH}
& \textbf{SVH}
\\

\midrule

Base
& $1.0\times$ & $1.0\times$ & $1.0\times$
& $1.0\times$ & $1.0\times$ & $1.0\times$
& $1.0\times$ & $1.0\times$ & $1.0\times$
\\

\midrule

VCD~\citep{VCD_2024_CVPR}
& $2.2\times$ & $3.0\times$ & $1.6\times$
& $2.9\times$ & $3.2\times$ & $2.9\times$
& $3.0\times$ & $3.3\times$ & $3.0\times$
\\

ICD~\citep{ICD_2024_FindingsACL}
& $2.1\times$ & $2.8\times$ & $1.6\times$
& $3.8\times$ & $4.1\times$ & $3.0\times$
& $3.7\times$ & $4.1\times$ & $3.2\times$
\\

CODE~\citep{CODE_2024_NeurIPS}
& $2.7\times$ & $3.8\times$ & $2.0\times$
& $3.6\times$ & $3.8\times$ & $2.1\times$
& $4.0\times$ & $3.8\times$ & $2.2\times$
\\

SID~\citep{SID_2025_ICLR}
& \textbf{$1.2\times$}
& \textbf{$1.4\times$}
& \textbf{$1.1\times$}
& $1.5\times$
& $1.5\times$
& \textbf{$1.5\times$}
& \textbf{$1.4\times$}
& \textbf{$1.4\times$}
& \textbf{$1.5\times$}
\\

AVCD~\citep{AVCD_2025_NeurIPS}
& $1.3\times$ & $1.6\times$ & $1.2\times$
& $2.6\times$ & $2.7\times$ & $3.9\times$
& $1.7\times$ & $2.1\times$ & $5.7\times$
\\

ASD~\citep{ASD_2026_ICML}
& $1.9\times$ & $2.4\times$ & $1.7\times$
& $3.6\times$ & $3.5\times$ & $5.3\times$
& $3.3\times$ & $3.3\times$ & $5.9\times$
\\

MAD~\citep{MAD_2026_CVPR}
& $2.4\times$ & $2.7\times$ & $1.4\times$
& $3.0\times$ & $2.9\times$ & $2.3\times$
& $2.9\times$ & $2.8\times$ & $2.4\times$
\\

\rowcolor{ourslavender}
\textbf{RED (Ours)}
& $1.6\times$
& \textbf{$1.4\times$}
& $1.2\times$
& \textbf{$1.4\times$}
& \textbf{$1.4\times$}
& $1.7\times$
& $1.7\times$
& \textbf{$1.4\times$}
& $1.7\times$
\\

\bottomrule
\end{tabularx}
\end{table*}

\paragraph{Accuracy--TTFT trade-off under SID pruning.}
  Because SID~\citep{SID_2025_ICLR} can reduce inference cost
  by pruning perceptual tokens, we compare its accuracy--TTFT
  trade-off under a common 2-fps configuration.
  As shown in Table~\ref{tab:sid_pruning_comparison},
  more aggressive pruning reduces SID's TTFT but also lowers
  its average CMM accuracy.
  With 90\% of audio and visual tokens removed, SID reaches
  81.54\% accuracy at $1.509\times$ Base TTFT, whereas \ours
  reaches 85.00\% at $1.433\times$ under the same configuration.

\begin{table*}[t]
\caption{
\textbf{Accuracy--TTFT trade-off of SID~\citep{SID_2025_ICLR} pruning on CMM.}
  Results with Qwen2.5-Omni-3B under a common 2-fps configuration.
  Accuracy (\%) is reported on all 2,400 questions, and TTFT is normalized to the paired Base.
  \textbf{Bold} indicates the best accuracy or lowest non-base TTFT.
}
\label{tab:sid_pruning_comparison}
\vspace{-.4em}
\centering
\footnotesize
\setlength{\tabcolsep}{0.85pt}
\renewcommand{\arraystretch}{1.04}
\begin{tabularx}{\textwidth}{@{}l*{6}{>{\centering\arraybackslash}X}>{\columncolor{avggray}\centering\arraybackslash}X*{2}{>{\centering\arraybackslash}X}@{}}
\toprule
\textbf{Method} & \textbf{VL} & \textbf{AL} & \textbf{VAL} & \textbf{Vis.} & \textbf{Aud.} & \textbf{Lang.} & \textbf{Avg.} & \textbf{TTFT (s)} & \textbf{Relative} \\
\midrule
Base model & \textbf{90.50} & \textbf{84.75} & 89.50 & 69.25 & 84.50 & 75.25 & 82.29 & 2.717 & $1.000\times$ \\
\midrule
SID (0\% removed) & \textbf{90.50} & \textbf{84.75} & 89.50 & 69.25 & 84.50 & 75.25 & 82.29 & 4.379 & $1.612\times$ \\
SID (60\% removed) & 87.50 & 83.75 & 88.25 & 71.25 & 81.25 & 77.75 & 81.63 & 4.179 & $1.538\times$ \\
SID (90\% removed) & 89.25 & 84.25 & 91.00 & 71.25 & 74.50 & \textbf{79.00} & 81.54 & 4.101 & $1.509\times$ \\
\rowcolor{ourslavender}
\textbf{RED (Ours)} & 88.25 & \textbf{84.75} & \textbf{95.00} & \textbf{76.25} & \textbf{88.50} & 77.25 & \cellcolor{redavggray}\textbf{85.00} & \textbf{3.893} & $\mathbf{1.433}\times$ \\
\bottomrule
\end{tabularx}
\end{table*}

\section{Experimental details}
\label{app:implementation}

\subsection{Evaluation protocol}
\label{app:protocol}

\paragraph{Evaluation.}
The results in Table~\ref{tab:main_full} are our reproductions using frozen backbones and deterministic decoding.
All methods are evaluated on the complete CMM, AVHBench, and SVHalluc benchmarks, with the unweighted mean reported over the six CMM subsets, three AVHBench tasks, and six SVHalluc tasks, respectively.
For MUSIC-AVQA, we use all 9,129 questions in the official corrected test annotations; MUSIC-AVQA-Hard contains 2,508 questions.
Method-specific decoding settings follow the original papers or released implementations unless stated otherwise.

\paragraph{Parameters and sampling.}
AVCD~\citep{AVCD_2025_NeurIPS} and MAD~\citep{MAD_2026_CVPR} use 100 randomly sampled labeled examples from each dataset to determine their decoding strengths.
\ours follows the same 100-example setting and randomly samples 100 questions from each benchmark to estimate its score statistics without using answer labels.
The resulting statistics are used throughout evaluation on the corresponding complete benchmark.
For the remaining baselines, we follow the decoding settings reported in the corresponding papers or released implementations.

\subsection{Baselines}
\label{app:baselines}
We compare RED with two groups of training-free decoding methods:
(1) audio-visual hallucination methods, including
AVCD~\citep{AVCD_2025_NeurIPS},
ASD~\citep{ASD_2026_ICML}, and
MAD~\citep{MAD_2026_CVPR}; and
(2) vision-language hallucination methods, including
VCD~\citep{VCD_2024_CVPR},
ICD~\citep{ICD_2024_FindingsACL},
CODE~\citep{CODE_2024_NeurIPS}, and
SID~\citep{SID_2025_ICLR}.
For the second group, we follow the trimodal extension protocol of
AVCD~\citep{AVCD_2025_NeurIPS}, adapting each method to audio, video, and joint audio-visual inputs while preserving its original decoding rule. For ASD~\citep{ASD_2026_ICML}, originally evaluated for caption generation,
we adapt its attention intervention to binary QA as detailed below.

\paragraph{Base model.}
The pretrained backbones (\ie, VideoLLaMA2.1-7B-AV, Qwen2.5-Omni-3B, and Qwen2.5-Omni-7B) are evaluated using their standard decoding procedures without fine-tuning or additional correction.

\paragrapht{VCD~\citep{VCD_2024_CVPR}} contrasts the original input with a corrupted counterpart.
Following the audio-visual extension introduced in AVCD~\citep{AVCD_2025_NeurIPS}, we construct audio, video, and joint corruption branches while retaining the original contrastive decoding rule.

\paragrapht{ICD~\citep{ICD_2024_FindingsACL}} obtains a contrastive prediction from a disturbance instruction.
We extend it to the audio-visual setting using audio-focused, video-focused, and joint disturbance branches while keeping the corresponding media inputs unchanged.

\paragrapht{CODE~\citep{CODE_2024_NeurIPS}} contrasts the original prediction with one conditioned on a self-generated description.
For the audio-visual setting, we generate contrastive descriptions covering both audio and video and otherwise retain the released decoding rule.

\paragrapht{SID~\citep{SID_2025_ICLR}} constructs a contrastive branch from low-importance perceptual tokens.
Following the audio-visual extension used in AVCD~\citep{AVCD_2025_NeurIPS}, we select low-importance tokens separately from the audio and video streams while retaining the original decoding rule.

\paragrapht{AVCD~\citep{AVCD_2025_NeurIPS}} identifies modality dominance and constructs attention-masked contrastive branches accordingly.
We follow its original audio-visual formulation and adapt only the modality-token indexing and attention hooks required by each backbone architecture.

\paragraph{ASD~\citep{ASD_2026_ICML}.}
ASD shifts attention from unimodal toward cross-modal sink tokens and combines the original and recalibrated predictions with an adaptive weight.
We retain the original intervention and apply it during answer generation for binary QA.
For VideoLLaMA2.1-7B-AV, sink tokens are identified separately within the audio and video streams, with the corresponding dimensions and threshold re-estimated from its activations.
We use the official implementation and authors' hyperparameters without additional QA-specific tuning.
On Qwen2.5-Omni-7B, the adaptive gate affects 37\% of examples and changes 15\% of predictions; setting $\alpha=0$ exactly recovers the base outputs.

\paragrapht{MAD~\citep{MAD_2026_CVPR}} estimates which modalities are required by the question and uses the resulting relevance scores to combine modality-conditioned predictions.
We follow its original audio-visual modality assessment and adaptive decoding rule for each backbone.

\subsection{Benchmark details and metrics}
\label{sec:benchmark_subsets}
\label{app:benchmark}

\paragraph{CMM.}
The Curse of Multi-Modalities (CMM)~\citep{CMM_2025_NeurIPS} evaluates hallucinations across language, visual, and audio modalities through binary object- and event-level probing.
It contains six subsets organized by two hallucination sources.
Visual-Language (VL), Audio-Language (AL), and Visual-Audio-Language (VAL) evaluate spurious inter-modality correlations, while Visual Dominance (Vis.), Audio Dominance (Aud.), and Language Dominance (Lang.) evaluate overreliance on unimodal priors.
Each subset contains 200 samples, and each sample is paired with one positive and one negative probing question, resulting in 1,200 samples and 2,400 Yes/No questions in total.
We evaluate the complete benchmark and report the unweighted mean accuracy over the six subsets.
For compactness, Table~\ref{tab:main_full} reports the three modality-dominance subsets as Vis., Aud., and Lang., together with the six-subset average; full per-subset results are provided in Table~\ref{tab:cmm_detailed}.

\paragraph{AVHBench.}
AVHBench~\citep{AVHBench_2025_ICLR} evaluates cross-modal hallucinations in audio-visual understanding.
The full benchmark contains three binary judgment tasks and an audio-visual captioning task.
We evaluate the three judgment tasks: Audio-driven Video Hallucination (ADV), Video-driven Audio Hallucination (VDA), and Audio-Visual Matching (AVM).
ADV tests whether audio induces hallucinated visual content, VDA tests whether visual content induces hallucinated audio, and AVM evaluates whether the audio and visual streams correspond.
The released evaluation set contains 5,302 binary judgment questions over 2,136 videos.
We report the unweighted mean accuracy over the three judgment tasks.

\paragraph{SVHalluc.}
SVHalluc~\citep{SVHalluc_2026_CVPR} evaluates speech--vision hallucinations along semantic and temporal dimensions.
The semantic tasks are Global Semantic Alignment (GSA), Fine-Grained Semantic Alignment (FGSA), and Cross-Modal Semantic Binding (CMSB).
The temporal tasks are Temporal Alignment (TA), Temporal Forecasting (TF), and Cross-Modal Temporal Binding (CMTB).
GSA, FGSA, CMSB, and TA use two-choice questions, while TF and CMTB use multiple-choice questions.
SVHalluc is constructed from the validation split of YouCook2 and contains 2,405 video-question pairs: 444 GSA, 444 FGSA, 534 CMSB, 368 TA, 276 TF, and 339 CMTB questions.
All models are evaluated in a zero-shot setting without additional training.
In Table~\ref{tab:main_full}, Sem. and Temp. denote the unweighted means over the three semantic and three temporal tasks, respectively, and Avg. denotes the unweighted mean over all six tasks.
Full task-level results are provided in Table~\ref{tab:svhalluc_detailed}.

\paragraph{MUSIC-AVQA.}
MUSIC-AVQA~\citep{MUSIC_AVQA_2022_CVPR} evaluates general audio-visual question answering in musical performance videos.
The original release contains 9,288 videos spanning 22 instrument categories and more than 150 hours of video, with 45,867 question-answer pairs constructed from 33 question templates.
The benchmark covers nine question types across audio, visual, and joint audio-visual settings, including counting, comparison, location, existence, and temporal reasoning.
MUSIC-AVQA is formulated as open-ended question answering with 42 possible answer categories.
The original dataset contains 32,087 training, 4,595 validation, and 9,185 test question-answer pairs.
We use the official corrected test annotations, which contain 9,129 questions, and report overall accuracy on this complete corrected split.\footnote{Official corrected annotations: \url{https://github.com/GeWu-Lab/MUSIC-AVQA/blob/main/data/json_update/avqa-test.json}.}
The evidence-selection distribution in Figure~\ref{fig:contribution_selection_distribution} uses the same 9,129 questions.

\paragraph{MUSIC-AVQA-Hard.}
MUSIC-AVQA-Hard~\citep{Kim_2026_Interspeech} is an audio-sensitive subset of the MUSIC-AVQA test split.
It is constructed using a single-frame shortcut test in which GPT-4o receives only the temporally central video frame, without audio or additional frames, in two independent runs.
Questions answered correctly in both runs are removed, leaving examples that are less susceptible to vision-only shortcuts.
The subset preserves the original MUSIC-AVQA question and answer format.
Our evaluation uses 2,508 questions from this filtered subset and reports overall accuracy.
All 2,508 questions are included in the evidence-selection distribution.

\paragraph{Metrics.}
We report accuracy as the primary metric across all benchmarks.
For CMM, Avg. is the unweighted mean over all six subsets; Table~\ref{tab:main_full} also reports the three modality-dominance subsets individually.
For AVHBench, Avg. is the unweighted mean over its three judgment tasks.
For SVHalluc, Sem. and Temp. are the unweighted means over the three semantic and three temporal tasks, respectively, and Avg. is the unweighted mean over all six tasks.
For MUSIC-AVQA and MUSIC-AVQA-Hard, we report accuracy over all 9,129 and 2,508 questions in the corresponding evaluation splits, respectively.
For analyses based on continuous binary answer scores, we additionally report AUROC to measure discrimination independently of a fixed decision threshold.

\subsection{Evidence-intervention protocol and results}
\label{app:intervention_details}

A correct prediction does not necessarily imply that the model relied on the evidence required by the question~\citep{Xiao_2024_CVPR,MIRAGEPerf_2025_NeurIPS}.
Following intervention-based diagnostics~\citep{Feder_2022_TACL,Ribeiro_2020_ACL,Kaushik_2020_ICLR,gardner2020contrast}, we evaluate 3{,}359 retained intervention pairs from AVHBench~\citep{AVHBench_2025_ICLR}: 1{,}621 answer-changing pairs and 1{,}738 answer-preserving pairs.
Sensitivity measures the fraction of answer-changing pairs for which both the original and intervened inputs are answered correctly, whereas invariance violation measures prediction changes under answer-preserving non-required-modality swaps.
Within each test, all methods use the same retained pairs and matched donor assignments.

\paragraph{Annotation-guided pair construction.}
We construct original--contrastive pairs from the Video-driven Audio Hallucination and Audio-driven Video Hallucination tasks in AVHBench~\citep{AVHBench_2025_ICLR}, whose questions require audio and visual evidence, respectively.
Each pair retains the question and replaces one modality with the corresponding stream from a donor clip.
For sensitivity, we select donors with explicit opposite-label annotations for the same queried entity and required modality.
A Yes-to-No replacement is supported by the donor's No annotation, and a No-to-Yes replacement by its Yes annotation.
We retain the question and the original stream of the other modality and assign the donor's annotated answer to the replacement input.
For invariance, we retain the question and the required source modality while replacing the non-required modality: video for audio-required questions and audio for video-required questions.
The answer remains unchanged because its determining evidence is retained.
The evaluation comprises 1{,}621 sensitivity pairs and 1{,}738 invariance pairs from 2{,}680 distinct source video--question items (Table~\ref{tab:intervention_construction}).
The sensitivity set contains 804/524 Yes-to-No/No-to-Yes pairs for audio-required questions and 192/101 for video-required questions; the invariance set contains 1{,}064 and 674 pairs, respectively.

\begin{table}[t]
\centering
\captionof{table}{
\textbf{Construction of evidence interventions.}
Required-modality swaps change the answer, whereas irrelevant-modality
swaps preserve it.
Pairs denote retained interventions and Samples denote unique source
video--question items; the total is deduplicated.
}
\label{tab:intervention_construction}

\vspace{-0.35em}

\centering
\begin{minipage}{0.70\linewidth}
\centering

\begingroup
\footnotesize
\setlength{\tabcolsep}{1.5pt}
\renewcommand{\arraystretch}{1.05}

\begin{tabularx}{\linewidth}{
@{}
>{\raggedright\arraybackslash}p{0.08\linewidth}
>{\centering\arraybackslash}p{0.12\linewidth}
>{\centering\arraybackslash}p{0.07\linewidth}
>{\centering\arraybackslash}p{0.20\linewidth}
>{\centering\arraybackslash}p{0.18\linewidth}
>{\raggedleft\arraybackslash}p{0.12\linewidth}
>{\raggedleft\arraybackslash}X
@{}
}
\toprule
\textbf{Test} & \textbf{Req.} & \textbf{Swap} & \textbf{Gold}
& \textbf{Donor} & \textbf{Pairs} & \textbf{Samples} \\
\midrule

\multirow{4}{*}{Sens.}
& \multirow{2}{*}{Audio}
& \multirow{2}{*}{A}
& Yes$\rightarrow$No
& Annotated
& 804
& 804
\\

&
&
& No$\rightarrow$Yes
& Annotated
& 524
& 524
\\

\cmidrule(lr){2-7}

&
\multirow{2}{*}{Video}
& \multirow{2}{*}{V}
& Yes$\rightarrow$No
& Annotated
& 192
& 192
\\

&
&
& No$\rightarrow$Yes
& Annotated
& 101
& 101
\\

\midrule

\multirow{2}{*}{Inv.}
& Video
& A
& unchanged
& Random
& 674
& 674
\\

& Audio
& V
& unchanged
& Random
& 1,064
& 1,064
\\

\midrule

\multicolumn{5}{@{}l}{Total (unique source samples)}
& 3,359
& 2,680
\\

\bottomrule
\end{tabularx}

\endgroup
\end{minipage}

\end{table}

\paragraph{Annotation and media verification.}
We cross-reference the original and donor QA records, replacement-media identifiers, and saved predictions to verify the queried entity, answer transition, and stream provenance.
Video packet payloads are compared with the expected original or donor source, and the required-modality source paths and donor assignments are checked across evaluation records.
For video replacements, we check the source and sample count for the retained, re-encoded audio.
We further examine the original and replacement evidence in sensitivity pairs using PANNs~\citep{PANNs_2020_TASLP} for audio and OWLv2~\citep{OWLv2_2023_NeurIPS} for sampled video frames.
We use the resulting event and object scores to check consistency with the queried entity and the annotation-defined answer transition. This provides complementary checks and identifies evidence that requires review.
AVHBench annotations determine the reference answers.
All methods are compared on the same retained pairs with matched donor assignments. For invariance, we use pairs with valid binary responses across methods and backbones.

\begin{figure*}[t]
\centering
\includegraphics[width=0.95\textwidth,keepaspectratio]{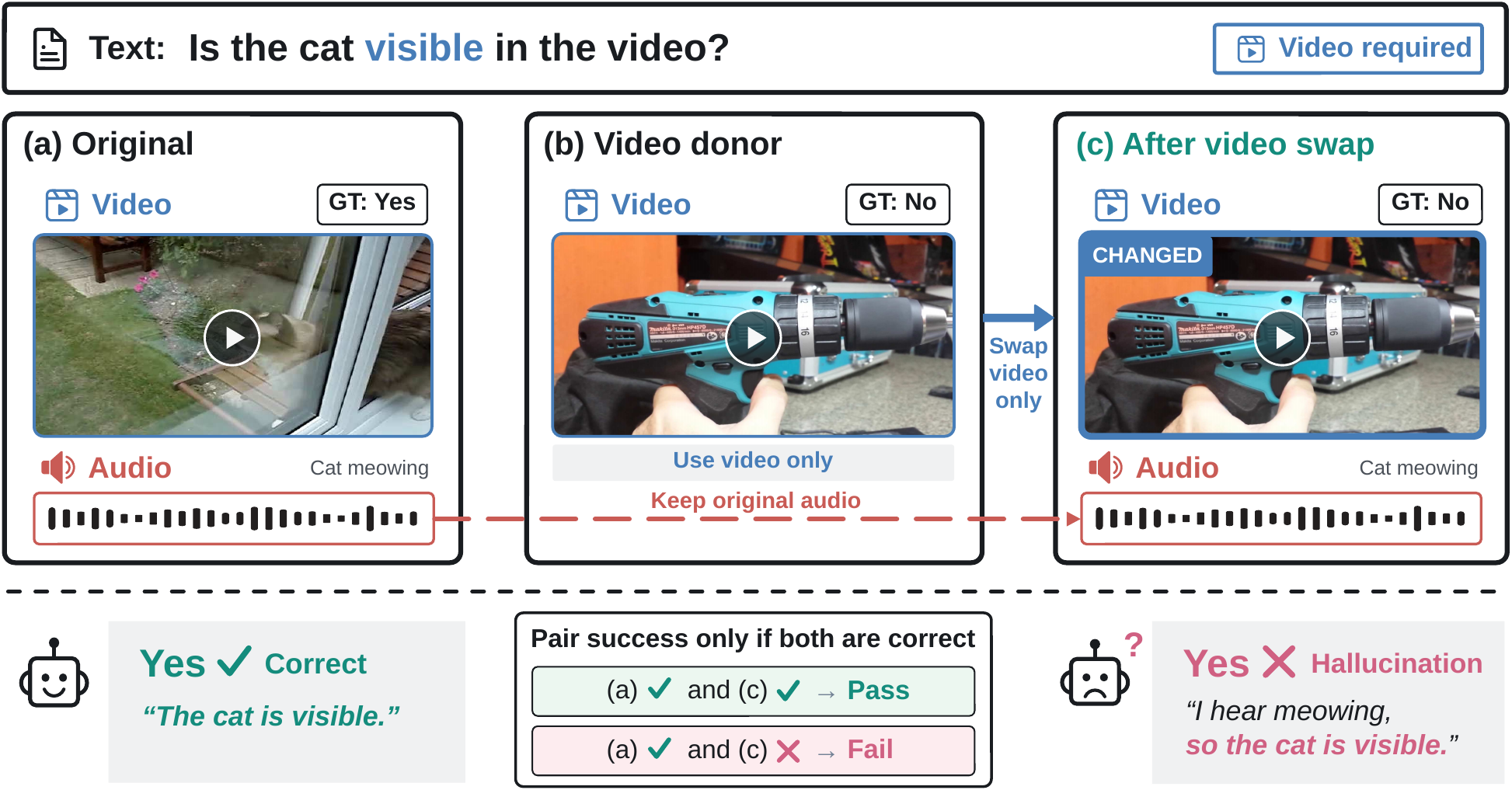}
\caption{
\textbf{Illustration of pair-level evidence intervention.}
For a video-required question, we replace the video of AVHBench sample 01161 with that of sample 00640 while retaining the original audio from 01161.
The reference answer changes accordingly, and the pair succeeds only when both the original and intervened inputs are answered correctly.
This is an illustrative pair, not a retained evaluation pair; waveforms are schematic.
}
\label{fig:intervention_example}
\end{figure*}

\begin{figure*}[t]
\centering
\includegraphics[width=\textwidth,keepaspectratio]{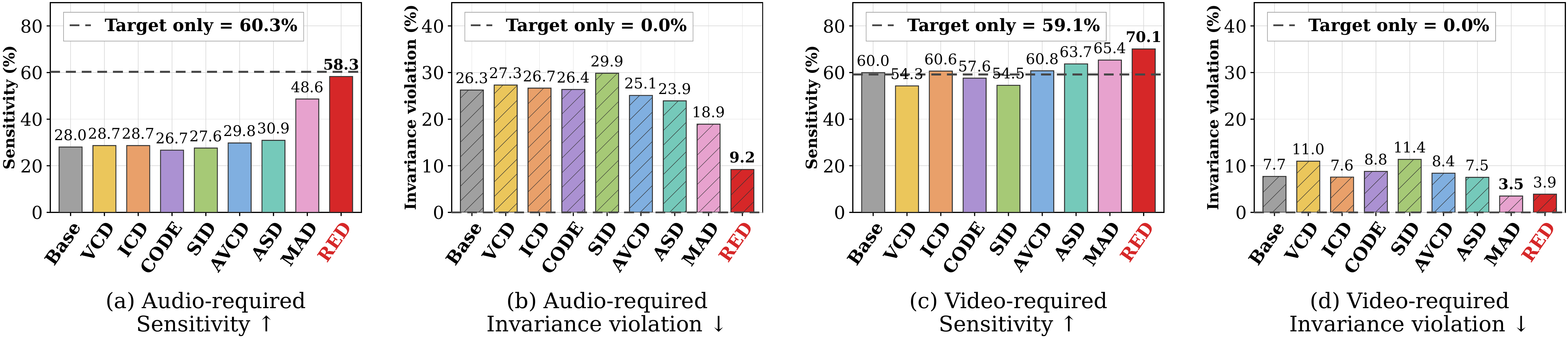}
\caption{\textbf{Evidence following under audio-visual interventions on AVHBench.} Sensitivity requires both answers in an answer-changing pair to be correct, while invariance violation measures prediction changes under non-required-modality swaps. Bars and whiskers denote the mean and min--max range across three backbones.}
\label{fig:sens_inv}
\end{figure*}

\paragraph{Pair-level evaluation.}
Sensitivity requires both the original and replacement answers to be correct.
Since their reference answers differ, consistently predicting either Yes or No cannot pass a pair: the model must answer the original correctly and update its answer according to the replacement evidence.
We average the Yes-to-No and No-to-Yes success rates equally so that both transition directions contribute equally despite their different sample counts; an invalid response counts as a failure.
Invariance violation is the fraction of answer-preserving pairs whose predictions change.
For the Target-only reference, only the question-required modality is provided.
Replacing a non-required modality therefore leaves the input unchanged, so the same prediction applies to both members, yielding zero invariance violation by construction.
Figure~\ref{fig:intervention_example} illustrates the pair-level sensitivity criterion: after replacing the required evidence while retaining the other modality, both the original and intervened inputs must be answered correctly.
Figure~\ref{fig:sens_inv} reports the modality-specific results, while Figure~\ref{fig:evidence_following} reports their corresponding averages over the two required modalities.

\paragraph{Evidence following under intervention.}
Figure~\ref{fig:sens_inv} examines whether the accuracy gains are accompanied by stronger dependence on question-required evidence.
On audio-required questions, \ours increases sensitivity from 28.0\% for Base to 58.3\%, while reducing invariance violation from 26.3\% to 9.2\%.
On video-required questions, sensitivity increases from 60.0\% to 70.1\%, while invariance violation decreases from 7.7\% to 3.9\%.
Among methods using both modalities, \ours achieves the highest sensitivity for both evidence types and the lowest invariance violation on audio-required questions.
For video-required questions, MAD~\citep{MAD_2026_CVPR} achieves a slightly lower invariance violation than \ours (3.5\% vs.\ 3.9\%).

\section{Additional experiments}
\label{app:additional}

\subsection{Detailed CMM and SVHalluc results}
\label{sec:detailed_hallucination_results}

Tables~\ref{tab:cmm_detailed} and~\ref{tab:svhalluc_detailed} provide subset- and task-level results corresponding to the aggregate CMM and SVHalluc results reported in the main text.
\ours achieves the highest average accuracy across all three backbones on both benchmarks.

\begin{table*}[t]
\caption{\textbf{Detailed performance on CMM.}
Accuracy (\%) on all six CMM subsets; higher is better.
VL, AL, and VAL denote Visual-Language, Audio-Language, and Visual-Audio-Language, respectively.
Vis., Aud., and Lang.\ denote Visual Dominance, Audio Dominance, and Language Dominance, respectively.
Avg.\ is the unweighted mean over all six subsets.
\textbf{Bold} indicates the best result within each backbone.}
\label{tab:cmm_detailed}
\vspace{-.4em}
\centering
\footnotesize
\setlength{\tabcolsep}{0.85pt}
\renewcommand{\arraystretch}{1.04}
\begin{tabularx}{\textwidth}{@{}l*{6}{>{\centering\arraybackslash}X}>{\columncolor{avggray}\centering\arraybackslash}X@{}}
\toprule
\textbf{Method} & \textbf{VL} & \textbf{AL} & \textbf{VAL} & \textbf{Vis.} & \textbf{Aud.} & \textbf{Lang.} & \textbf{Avg.} \\
\midrule
\multicolumn{8}{c}{\textbf{VideoLLaMA2.1-7B-AV}} \\
\cmidrule(lr){1-8}
Base model & \textbf{93.5} & 81.2 & 92.2 & 73.5 & 84.0 & 72.5 & 82.8 \\
VCD~\citep{VCD_2024_CVPR} & 92.5 & 77.5 & 89.8 & 70.8 & 84.5 & 73.8 & 81.5 \\
ICD~\citep{ICD_2024_FindingsACL} & 93.0 & 81.0 & 91.5 & 73.5 & 84.2 & 72.5 & 82.6 \\
CODE~\citep{CODE_2024_NeurIPS} & 92.8 & 81.8 & 91.2 & 72.5 & 84.0 & 72.0 & 82.4 \\
SID~\citep{SID_2025_ICLR} & 90.2 & 75.0 & 86.2 & 65.8 & 83.5 & 73.0 & 79.0 \\
AVCD~\citep{AVCD_2025_NeurIPS} & 93.2 & 77.8 & 91.5 & 73.5 & 82.5 & 70.8 & 81.5 \\
ASD~\citep{ASD_2026_ICML} & 92.5 & 78.8 & 91.2 & 74.5 & 85.5 & 72.0 & 82.4 \\
MAD~\citep{MAD_2026_CVPR} & 92.5 & 80.8 & \textbf{92.5} & 82.0 & 82.8 & \textbf{77.8} & 84.7 \\
\rowcolor{ourslavender}
\textbf{RED (Ours)} & 91.8 & \textbf{83.2} & 91.0 & \textbf{83.5} & \textbf{88.0} & 76.2 & \cellcolor{redavggray}\textbf{85.6} \\
\midrule
\multicolumn{8}{c}{\textbf{Qwen2.5-Omni-3B}} \\
\cmidrule(lr){1-8}
Base model & 90.5 & \textbf{87.0} & 92.2 & 69.2 & 84.2 & 76.5 & 83.3 \\
VCD~\citep{VCD_2024_CVPR} & 90.0 & 85.5 & 93.0 & 73.2 & 82.0 & 76.2 & 83.3 \\
ICD~\citep{ICD_2024_FindingsACL} & 90.5 & 85.0 & 91.8 & 68.5 & 84.0 & 75.8 & 82.6 \\
CODE~\citep{CODE_2024_NeurIPS} & 91.5 & 84.5 & 92.8 & 69.5 & 85.2 & 77.2 & 83.5 \\
SID~\citep{SID_2025_ICLR} & 91.2 & 85.0 & 91.8 & 71.2 & 78.0 & 77.0 & 82.4 \\
AVCD~\citep{AVCD_2025_NeurIPS} & 89.5 & 85.0 & 92.2 & 70.0 & 85.0 & 75.5 & 82.9 \\
ASD~\citep{ASD_2026_ICML} & 91.5 & 86.5 & 92.8 & 70.0 & 84.8 & 78.0 & 83.9 \\
MAD~\citep{MAD_2026_CVPR} & 91.2 & 85.5 & 90.8 & 74.5 & \textbf{87.0} & 78.8 & 84.6 \\
\rowcolor{ourslavender}
\textbf{RED (Ours)} & \textbf{92.5} & 84.2 & \textbf{96.0} & \textbf{75.8} & \textbf{87.0} & \textbf{81.5} & \cellcolor{redavggray}\textbf{86.2} \\
\midrule
\multicolumn{8}{c}{\textbf{Qwen2.5-Omni-7B}} \\
\cmidrule(lr){1-8}
Base model & \textbf{94.8} & 83.0 & 92.5 & 69.5 & 71.5 & 82.2 & 82.2 \\
VCD~\citep{VCD_2024_CVPR} & 92.8 & 84.2 & 93.2 & 69.0 & 68.5 & 82.5 & 81.7 \\
ICD~\citep{ICD_2024_FindingsACL} & 93.5 & 84.2 & 92.5 & 68.8 & 71.2 & 81.8 & 82.0 \\
CODE~\citep{CODE_2024_NeurIPS} & 93.8 & 83.5 & 91.8 & 68.0 & 68.5 & 82.8 & 81.4 \\
SID~\citep{SID_2025_ICLR} & 92.0 & 86.5 & 91.0 & 73.5 & 65.8 & \textbf{83.5} & 82.0 \\
AVCD~\citep{AVCD_2025_NeurIPS} & 87.5 & 84.2 & 93.0 & 69.5 & 72.8 & 75.2 & 80.4 \\
ASD~\citep{ASD_2026_ICML} & 94.0 & 83.8 & 92.2 & 70.0 & 74.2 & 82.0 & 82.7 \\
MAD~\citep{MAD_2026_CVPR} & 93.8 & 85.0 & 94.5 & 77.0 & 83.2 & 83.0 & 86.1 \\
\rowcolor{ourslavender}
\textbf{RED (Ours)} & 93.5 & \textbf{88.5} & \textbf{95.2} & \textbf{86.0} & \textbf{88.8} & 83.0 & \cellcolor{redavggray}\textbf{89.2} \\
\bottomrule
\end{tabularx}
\end{table*}

The detailed CMM breakdown shows that the gains are concentrated on the modality-dominance subsets: \ours is best or tied best on Vis.\ and Aud.\ across all three backbones.
On Qwen2.5-Omni-7B, \ours achieves absolute gains of 16.5\% and 17.3\% over the Base model on Vis.\ and Aud., respectively, while the changes on VL, AL, and VAL are smaller.
This pattern further suggests that \ours is particularly effective when one modality can dominate the prediction.

\begin{table*}[t]
\caption{\textbf{Detailed performance on SVHalluc.}
Accuracy (\%) on all six SVHalluc tasks; higher is better.
GSA, FGSA, and CMSB denote Global Semantic Alignment, Fine-Grained Semantic Alignment, and Cross-Modal Semantic Binding, respectively.
TA, TF, and CMTB denote Temporal Alignment, Temporal Forecasting, and Cross-Modal Temporal Binding.
Avg.\ is the unweighted mean over all six tasks.
\textbf{Bold} indicates the best result within each backbone, including ties.}
\label{tab:svhalluc_detailed}
\vspace{-.4em}
\centering
\footnotesize
\setlength{\tabcolsep}{0.85pt}
\renewcommand{\arraystretch}{1.04}
\begin{tabularx}{\textwidth}{@{}l*{6}{>{\centering\arraybackslash}X}>{\columncolor{avggray}\centering\arraybackslash}X@{}}
\toprule
\multirow[c]{2}{*}{\raisebox{-1.5ex}{\textbf{Method}}} & \multicolumn{3}{c}{\textbf{Sem.}} & \multicolumn{3}{c}{\textbf{Temp.}} & \textbf{Avg.} \\
\cmidrule(lr){2-4}\cmidrule(lr){5-7}
& GSA & FGSA & CMSB & TA & TF & CMTB & \\
\midrule
\multicolumn{8}{c}{\textbf{VideoLLaMA2.1-7B-AV}} \\
\cmidrule(lr){1-8}
Base model & 50.0 & 67.8 & 71.3 & 50.0 & 30.4 & \textbf{50.1} & 53.3 \\
VCD~\citep{VCD_2024_CVPR} & 50.0 & 70.7 & 73.0 & 50.0 & 30.8 & \textbf{50.1} & 54.1 \\
ICD~\citep{ICD_2024_FindingsACL} & 50.0 & 68.2 & 71.3 & 50.0 & 30.1 & \textbf{50.1} & 53.3 \\
CODE~\citep{CODE_2024_NeurIPS} & 50.0 & 68.0 & 71.7 & 50.0 & 29.3 & \textbf{50.1} & 53.2 \\
SID~\citep{SID_2025_ICLR} & 50.0 & 66.2 & 71.9 & 50.0 & 29.3 & 49.6 & 52.8 \\
AVCD~\citep{AVCD_2025_NeurIPS} & 50.0 & 68.5 & 72.1 & 50.0 & 30.4 & 49.9 & 53.5 \\
ASD~\citep{ASD_2026_ICML} & 50.0 & 59.2 & 70.6 & 50.0 & 27.5 & 48.1 & 50.9 \\
MAD~\citep{MAD_2026_CVPR} & 50.0 & 75.7 & \textbf{78.8} & 50.0 & 31.9 & 49.3 & 55.9 \\
\rowcolor{ourslavender}
\textbf{RED (Ours)} & \textbf{51.4} & \textbf{81.5} & 75.1 & \textbf{52.7} & \textbf{33.0} & 48.7 & \cellcolor{redavggray}\textbf{57.1} \\
\midrule
\multicolumn{8}{c}{\textbf{Qwen2.5-Omni-3B}} \\
\cmidrule(lr){1-8}
Base model & \textbf{74.5} & 71.8 & 75.1 & 48.1 & 21.4 & 49.0 & 56.7 \\
VCD~\citep{VCD_2024_CVPR} & 74.1 & 70.0 & 74.2 & 48.6 & 24.6 & 51.3 & 57.2 \\
ICD~\citep{ICD_2024_FindingsACL} & 74.1 & 72.3 & 75.3 & 48.9 & 22.1 & 49.3 & 57.0 \\
CODE~\citep{CODE_2024_NeurIPS} & 71.6 & 70.5 & 74.2 & 49.2 & 23.6 & 49.0 & 56.3 \\
SID~\citep{SID_2025_ICLR} & 71.2 & 62.8 & 70.2 & 48.1 & 29.0 & 52.8 & 55.7 \\
AVCD~\citep{AVCD_2025_NeurIPS} & 73.6 & 71.8 & 75.5 & 50.0 & 26.4 & 47.8 & 57.5 \\
ASD~\citep{ASD_2026_ICML} & 74.3 & 72.3 & 74.9 & 49.2 & 23.2 & 49.0 & 57.1 \\
MAD~\citep{MAD_2026_CVPR} & 59.2 & \textbf{81.8} & 84.3 & \textbf{51.4} & \textbf{30.1} & 50.1 & 59.5 \\
\rowcolor{ourslavender}
\textbf{RED (Ours)} & 65.3 & 80.0 & \textbf{85.6} & 50.0 & 29.0 & \textbf{53.4} & \cellcolor{redavggray}\textbf{60.5} \\
\midrule
\multicolumn{8}{c}{\textbf{Qwen2.5-Omni-7B}} \\
\cmidrule(lr){1-8}
Base model & 69.6 & 83.3 & 75.5 & 52.2 & 33.7 & 48.7 & 60.5 \\
VCD~\citep{VCD_2024_CVPR} & 70.3 & 80.2 & 73.4 & \textbf{53.3} & 34.1 & 48.1 & 59.9 \\
ICD~\citep{ICD_2024_FindingsACL} & 69.8 & 82.4 & 75.5 & 51.9 & 33.7 & 49.0 & 60.4 \\
CODE~\citep{CODE_2024_NeurIPS} & 73.0 & 82.0 & 74.5 & 52.4 & \textbf{35.5} & 47.5 & 60.8 \\
SID~\citep{SID_2025_ICLR} & 70.3 & 78.4 & 70.4 & 53.0 & \textbf{35.5} & 50.1 & 59.6 \\
AVCD~\citep{AVCD_2025_NeurIPS} & 73.0 & 83.8 & 76.6 & 51.4 & 33.7 & \textbf{51.9} & 61.7 \\
ASD~\citep{ASD_2026_ICML} & 68.0 & 83.3 & 74.5 & 50.3 & 34.4 & 50.1 & 60.1 \\
MAD~\citep{MAD_2026_CVPR} & 68.2 & \textbf{89.6} & 86.3 & 51.6 & 33.7 & 49.9 & 63.2 \\
\rowcolor{ourslavender}
\textbf{RED (Ours)} & \textbf{77.0} & 81.3 & \textbf{87.3} & 52.4 & 33.3 & 49.0 & \cellcolor{redavggray}\textbf{63.4} \\
\bottomrule
\end{tabularx}
\end{table*}

The task-level breakdown shows that gains are uneven across SVHalluc: CMSB improves over the Base model for all three backbones, whereas temporal changes are more backbone-dependent.
For VideoLLaMA2.1-7B-AV, the near-chance GSA and TA results are consistent with the failure pattern reported in SVHalluc~\citep{SVHalluc_2026_CVPR}; for GSA, the benchmark attributes this behavior to an alignment bias toward positive correspondence.
Existing decoding baselines largely preserve this pattern.

\subsection{Evidence attenuation and recovery}
\label{sec:app_evidence_recovery}

\begin{figure*}[t]
\centering
\includegraphics[width=0.95\textwidth]{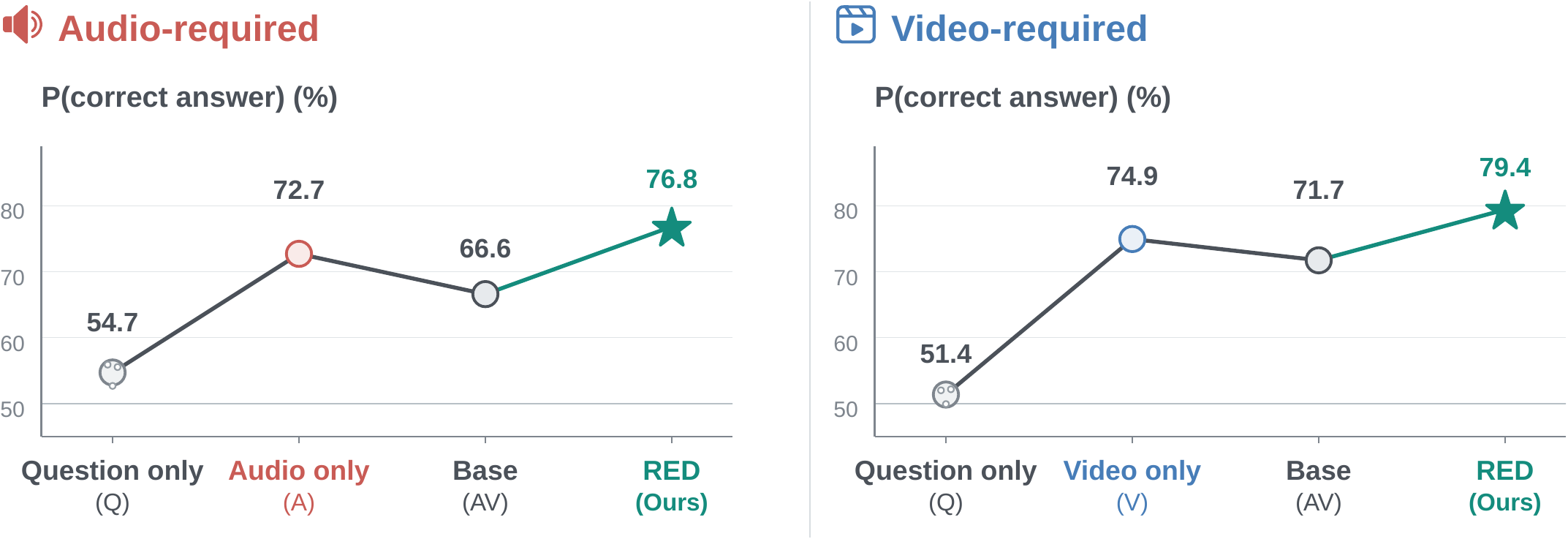}
\caption{
\textbf{Evidence attenuation and recovery across input conditions.}
Correct-answer probability on AVHBench for audio-required and video-required questions, averaged equally across Qwen2.5-Omni-3B, Qwen2.5-Omni-7B, and VideoLLaMA2.1-7B-AV.
The required modality provides substantially stronger support than question-only inference; \ours recovers and further strengthens this support under the full audio-visual input.
}
\label{fig:app_S12_answer_probability}
\end{figure*}

Figure~\ref{fig:app_S12_answer_probability} summarizes how support for the correct answer changes across question-only, required-modality-only, full audio-visual, and \ours predictions on AVHBench.
For both audio-required and video-required questions, providing the required modality substantially increases the correct-answer probability relative to question-only inference.
Under the full audio-visual input, part of this evidence gain is attenuated, while \ours recovers and further strengthens support for the correct answer while retaining both modalities.
Audio only and Video only denote the task-defined required-modality inputs.

\subsection{Comparison with PMI-based decoding}
\label{sec:pmi_decoding_comparison}

Table~\ref{tab:pmi_decoding_comparison} compares \ours with representative PMI-based decoding methods on CMM using Qwen2.5-Omni-3B.
\ours achieves the highest average accuracy among the compared methods.

\begin{table*}[t]
\caption{\textbf{Comparison with PMI-based decoding on CMM.} Accuracy (\%) with Qwen2.5-Omni-3B; higher is better.}
\label{tab:pmi_decoding_comparison}
\vspace{-.4em}
\centering
\footnotesize
\setlength{\tabcolsep}{0.85pt}
\renewcommand{\arraystretch}{1.04}
\begin{tabularx}{\textwidth}{@{}l*{6}{>{\centering\arraybackslash}X}>{\columncolor{avggray}\centering\arraybackslash}X@{}}
\toprule
\textbf{Method} & \textbf{VL} & \textbf{AL} & \textbf{VAL} & \textbf{Vis.} & \textbf{Aud.} & \textbf{Lang.} & \textbf{Avg.} \\
\midrule
Base model & 90.5 & \textbf{87.0} & 92.2 & 69.2 & 84.2 & 76.5 & 83.3 \\
\midrule
CAD~\citep{CAD_2024_NAACL} ($\alpha=0.5$) & 91.2 & 85.5 & 93.8 & 69.2 & 86.0 & 80.5 & 84.4 \\
M3ID~\citep{M3ID_2024_CVPR} & 91.2 & 86.2 & 93.8 & 69.5 & 86.0 & 80.5 & 84.5 \\
DCPMI~\citep{DynamicPMI_2026_FindingsACL} ($\beta=0.25$) & \textbf{93.0} & 85.5 & 94.2 & 70.2 & 86.5 & 77.2 & 84.5 \\
\rowcolor{ourslavender}
\textbf{RED (Ours)} & 92.5 & 84.2 & \textbf{96.0} & \textbf{75.8} & \textbf{87.0} & \textbf{81.5} & \cellcolor{redavggray}\textbf{86.2} \\
\bottomrule
\end{tabularx}
\end{table*}

\subsection{Detailed MUSIC-AVQA results}
\label{sec:music_detailed}

Tables~\ref{tab:music_avqa} and~\ref{tab:music_avqa_hard} provide question-type-level results corresponding to the aggregate MUSIC-AVQA results reported in the main text.
\ours reaches 67.1\% overall accuracy on the 9,129-question official corrected MUSIC-AVQA test split and 40.5\% on the 2,508-question MUSIC-AVQA-Hard subset.

\begin{table*}[t]
\caption{\textbf{Performance on MUSIC-AVQA.}
Accuracy (\%) with Qwen2.5-Omni-7B; higher is better.
Avg.\ denotes accuracy over all 9,129 questions in the official corrected test split.
\textbf{Bold} indicates the best displayed result in each column, including ties.
Cnt., Comp., Loc., Exist., and Temp. denote counting, comparative, location, existential, and temporal questions, respectively.}
\label{tab:music_avqa}
\vspace{-.4em}
\centering
\footnotesize
\setlength{\tabcolsep}{0.85pt}
\renewcommand{\arraystretch}{1.04}
\begin{tabularx}{\textwidth}{@{}l*{9}{>{\centering\arraybackslash}X}>{\columncolor{avggray}\centering\arraybackslash}X@{}}
\toprule
\multirow[c]{2}{*}{\raisebox{-1.5ex}{\textbf{Method}}} & \multicolumn{2}{c}{\textbf{Audio}} & \multicolumn{2}{c}{\textbf{Visual}} & \multicolumn{5}{c}{\textbf{Audio-Visual}} & \multicolumn{1}{c}{} \\
\cmidrule(lr){2-3}\cmidrule(lr){4-5}\cmidrule(lr){6-10}
& Cnt. & Comp. & Cnt. & Loc. & Exist. & Loc. & Cnt. & Comp. & Temp. & \textbf{Avg.} \\
\midrule
Base model & 78.6 & 59.3 & 71.9 & 84.2 & 75.5 & 47.4 & 54.2 & 57.0 & 43.1 & 64.5 \\
VCD~\citep{VCD_2024_CVPR} & 82.4 & 64.8 & 72.2 & 82.0 & 74.9 & 42.8 & 54.0 & 58.8 & 40.5 & 64.5 \\
ICD~\citep{ICD_2024_FindingsACL} & 82.3 & 63.6 & 72.2 & 84.1 & 75.6 & 44.6 & 56.6 & \textbf{58.9} & 44.3 & 65.7 \\
CODE~\citep{CODE_2024_NeurIPS} & 80.3 & 63.5 & 70.4 & 83.3 & 74.5 & 44.2 & 48.9 & \textbf{58.9} & 42.8 & 63.7 \\
SID~\citep{SID_2025_ICLR} & \textbf{84.0} & 61.1 & 72.9 & 80.2 & 72.9 & 42.9 & 51.1 & 54.9 & 36.9 & 62.9 \\
AVCD~\citep{AVCD_2025_NeurIPS} & 81.7 & 63.8 & 72.2 & 82.9 & 75.4 & 42.2 & 54.5 & 58.7 & 41.7 & 64.6 \\
ASD~\citep{ASD_2026_ICML} & 82.1 & 63.3 & 72.3 & 79.5 & 70.3 & 38.3 & 51.2 & 56.9 & 39.8 & 62.4 \\
MAD~\citep{MAD_2026_CVPR} & 83.6 & 64.5 & \textbf{73.0} & \textbf{86.8} & 66.4 & 50.4 & \textbf{62.6} & 57.9 & \textbf{49.4} & \textbf{67.1} \\
\rowcolor{ourslavender}
\textbf{RED (Ours)} & 79.7 & \textbf{65.2} & 68.1 & 82.1 & \textbf{80.1} & \textbf{60.5} & 57.2 & \textbf{58.9} & 47.3 & \cellcolor{redavggray}\textbf{67.1} \\
\bottomrule
\end{tabularx}
\end{table*}

\begin{table*}[t]
\caption{\textbf{Performance on MUSIC-AVQA-Hard.}
Accuracy (\%) with Qwen2.5-Omni-7B; higher is better.
Avg.\ denotes accuracy over all 2,508 questions.
\textbf{Bold} indicates the best displayed result in each column, including ties.
Cnt., Comp., Loc., Exist., and Temp. denote counting, comparative, location, existential, and temporal questions, respectively.}
\label{tab:music_avqa_hard}
\vspace{-.4em}
\centering
\footnotesize
\setlength{\tabcolsep}{0.85pt}
\renewcommand{\arraystretch}{1.04}
\begin{tabularx}{\textwidth}{@{}l*{7}{>{\centering\arraybackslash}X}>{\columncolor{avggray}\centering\arraybackslash}X@{}}
\toprule
\multirow[c]{2}{*}{\raisebox{-1.5ex}{\textbf{Method}}} & \multicolumn{2}{c}{\textbf{Audio}} & \multicolumn{5}{c}{\textbf{Audio-Visual}} & \multicolumn{1}{c}{} \\
\cmidrule(lr){2-3}\cmidrule(lr){4-8}
& Cnt. & Comp. & Exist. & Loc. & Cnt. & Comp. & Temp. & \textbf{Avg.} \\
\midrule
Base model & 59.7 & 39.9 & 43.3 & 19.6 & 32.9 & 31.5 & 17.5 & 31.9 \\
VCD~\citep{VCD_2024_CVPR} & \textbf{62.3} & 41.5 & 45.5 & 18.3 & 33.7 & 31.5 & 17.3 & 32.4 \\
ICD~\citep{ICD_2024_FindingsACL} & 60.2 & 39.9 & 43.8 & 19.1 & 35.6 & 31.8 & 19.1 & 32.7 \\
CODE~\citep{CODE_2024_NeurIPS} & 59.7 & 39.4 & 42.7 & 19.9 & 30.0 & 34.7 & 19.1 & 32.3 \\
SID~\citep{SID_2025_ICLR} & 59.7 & 39.4 & 49.7 & 19.1 & 27.5 & 32.8 & 14.7 & 31.5 \\
AVCD~\citep{AVCD_2025_NeurIPS} & 60.2 & 40.4 & 43.5 & 18.3 & 33.4 & 31.5 & 16.7 & 31.7 \\
ASD~\citep{ASD_2026_ICML} & 61.3 & 39.4 & 32.3 & 18.8 & 30.0 & 26.6 & 14.5 & 28.2 \\
MAD~\citep{MAD_2026_CVPR} & 60.7 & 44.6 & 41.0 & 22.5 & 41.1 & \textbf{37.7} & 22.8 & 36.0 \\
\rowcolor{ourslavender}
\textbf{RED (Ours)} & 47.1 & \textbf{50.3} & \textbf{55.3} & \textbf{41.6} & \textbf{42.1} & 36.7 & \textbf{23.9} & \cellcolor{redavggray}\textbf{40.5} \\
\bottomrule
\end{tabularx}
\end{table*}

\section{Qualitative analysis}
\label{app:qualitative}

\subsection{Evidence attenuation and recovery}
\label{sec:qualitative_evidence_recovery}

\begin{figure*}[t]
\centering
\begin{subfigure}[t]{\textwidth}
\centering
\includegraphics[width=\linewidth]{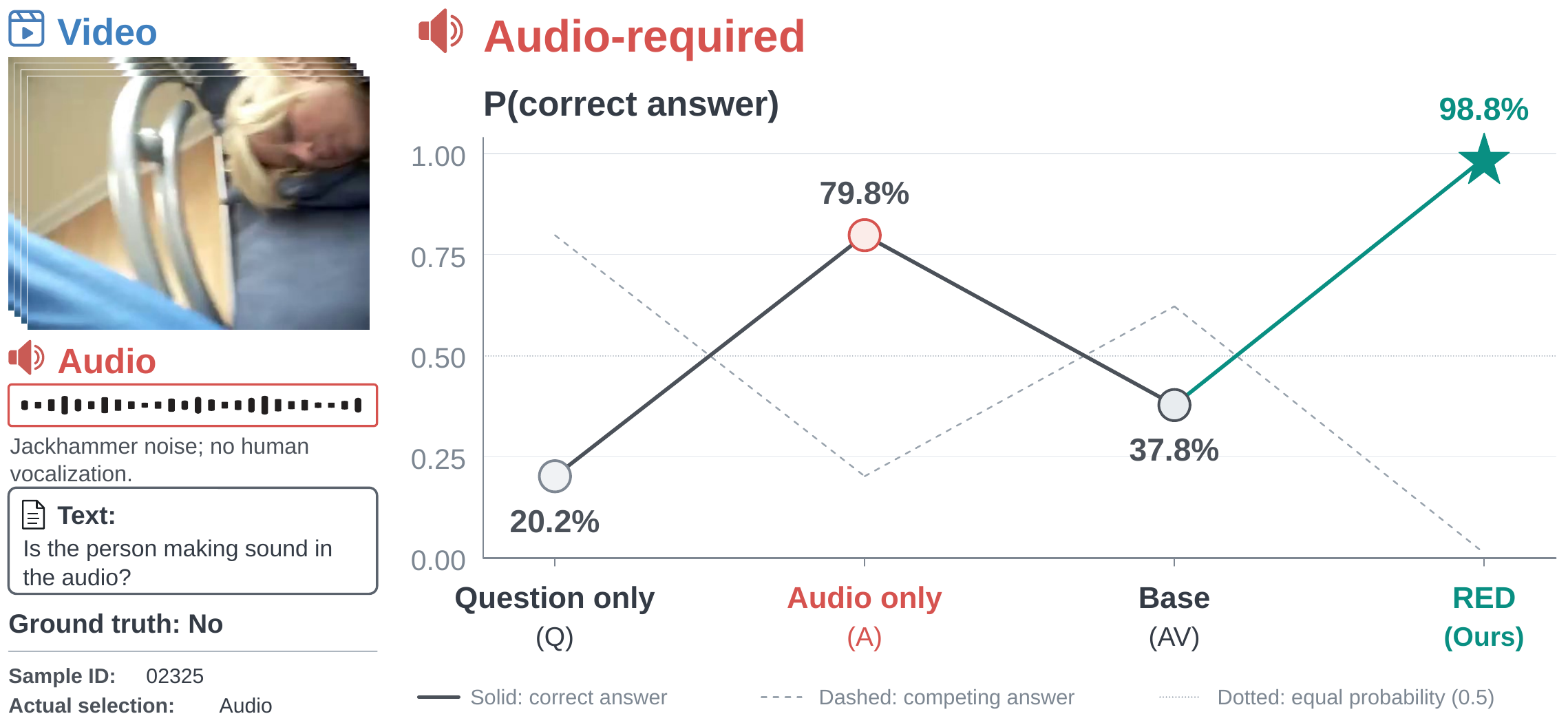}
\caption{Audio-required: human vocalization.}
\label{fig:attenuation_audio}
\end{subfigure}

\vspace{0.8em}

\begin{subfigure}[t]{\textwidth}
\centering
\includegraphics[width=\linewidth]{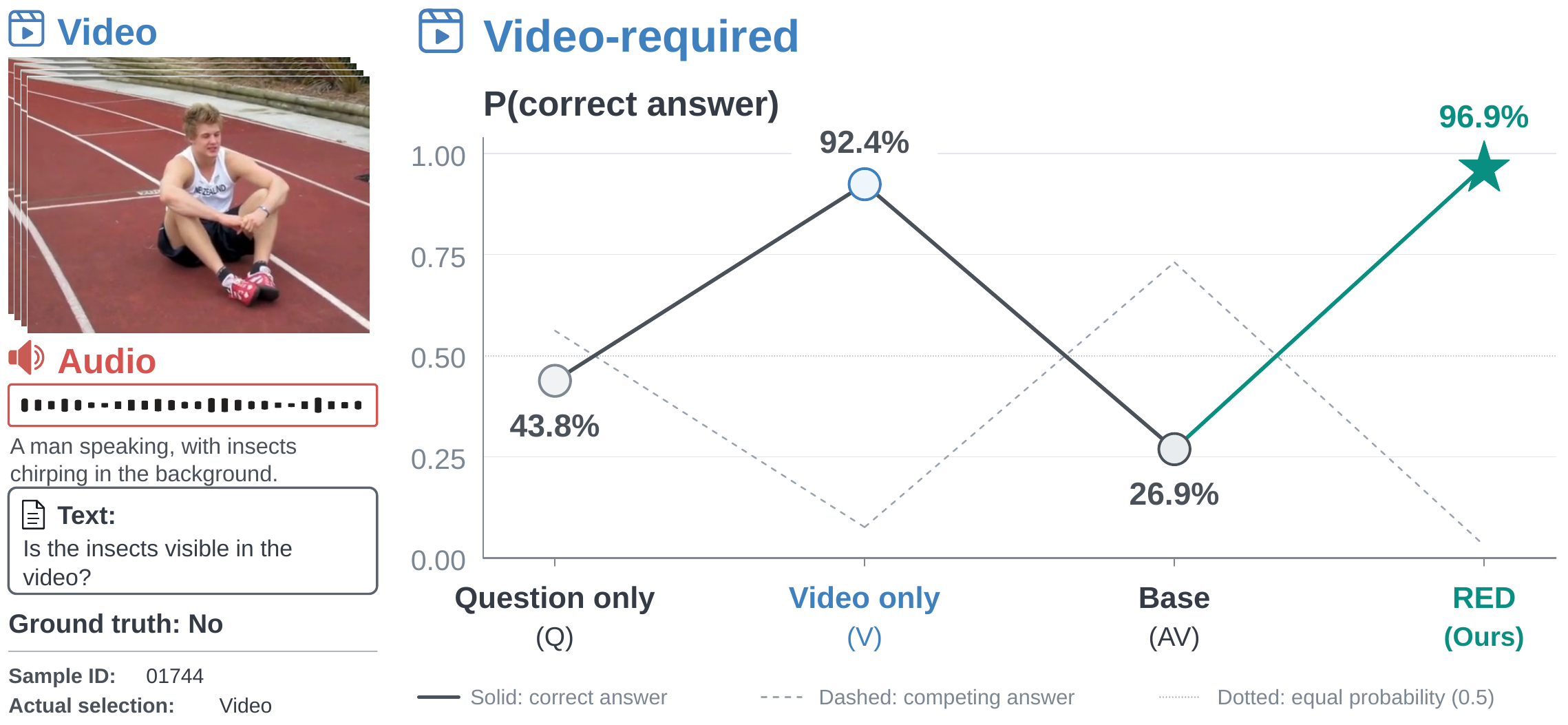}
\caption{Video-required: insect visibility.}
\label{fig:attenuation_video_1}
\end{subfigure}

\vspace{0.8em}

\begin{subfigure}[t]{\textwidth}
\centering
\includegraphics[width=\linewidth]{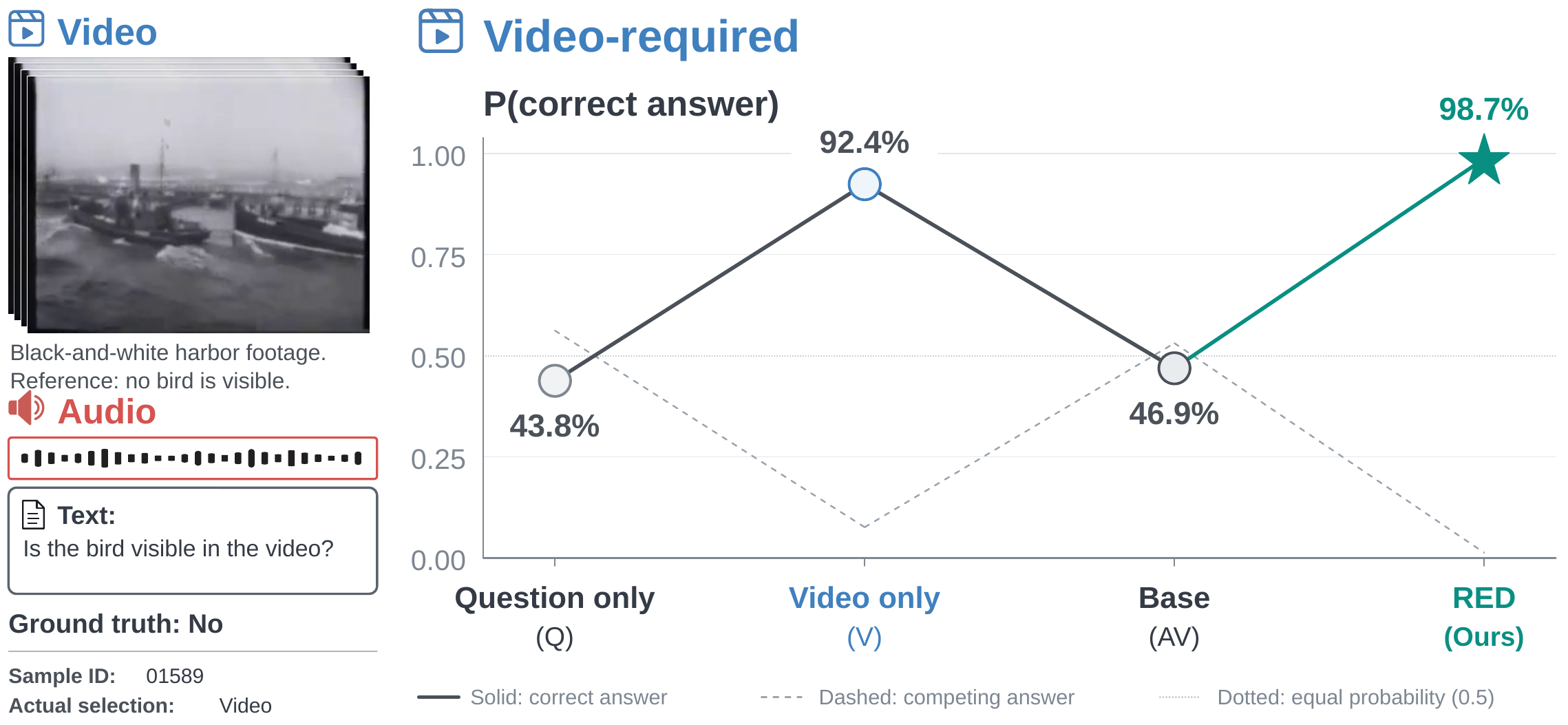}
\caption{Video-required: bird visibility.}
\label{fig:attenuation_video_2}
\end{subfigure}

\caption{
\textbf{Examples of evidence attenuation and recovery with Qwen2.5-Omni-7B.}
Representative AVHBench cases where \ours recovers question-required evidence under full audio-visual input.
}
\label{fig:app_red_qualitative}
\end{figure*}

Figure~\ref{fig:app_red_qualitative} provides individual examples of the evidence-attenuation pattern summarized in Figure~\ref{fig:app_S12_answer_probability}.
Across one audio-required and two video-required AVHBench cases, the required modality alone provides strong support for the correct answer, while \ours restores and strengthens this support under the full audio-visual input.
These examples illustrate how question-required evidence is recovered at the individual-example level.

\subsection{Representative hallucination cases}
\label{sec:qualitative_hallucination_cases}

\begin{figure*}[t]
\centering
\begin{subfigure}[t]{\textwidth}
\centering
\includegraphics[width=\linewidth]{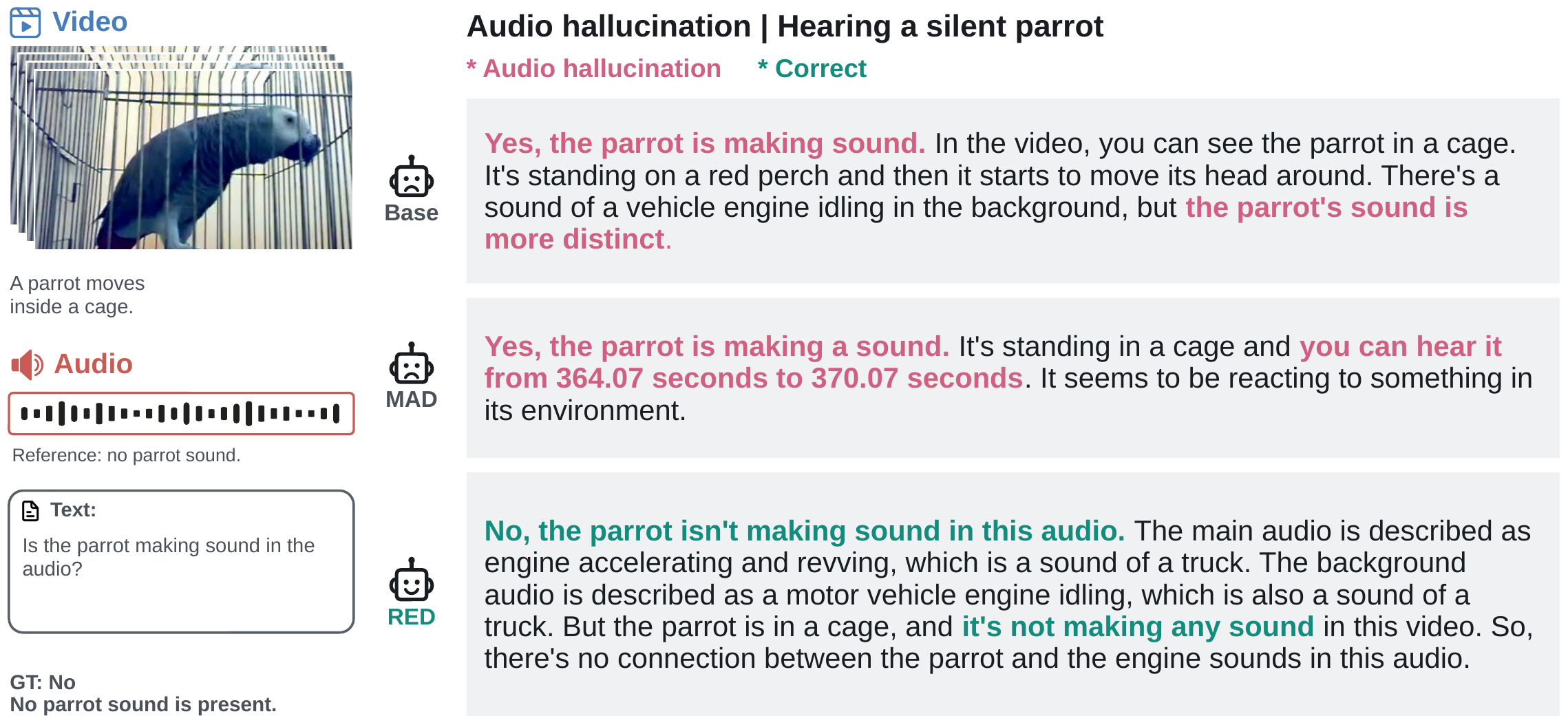}
\caption{Audio hallucination.}
\label{fig:qual_audio}
\end{subfigure}

\vspace{0.8em}

\begin{subfigure}[t]{\textwidth}
\centering
\includegraphics[width=\linewidth]{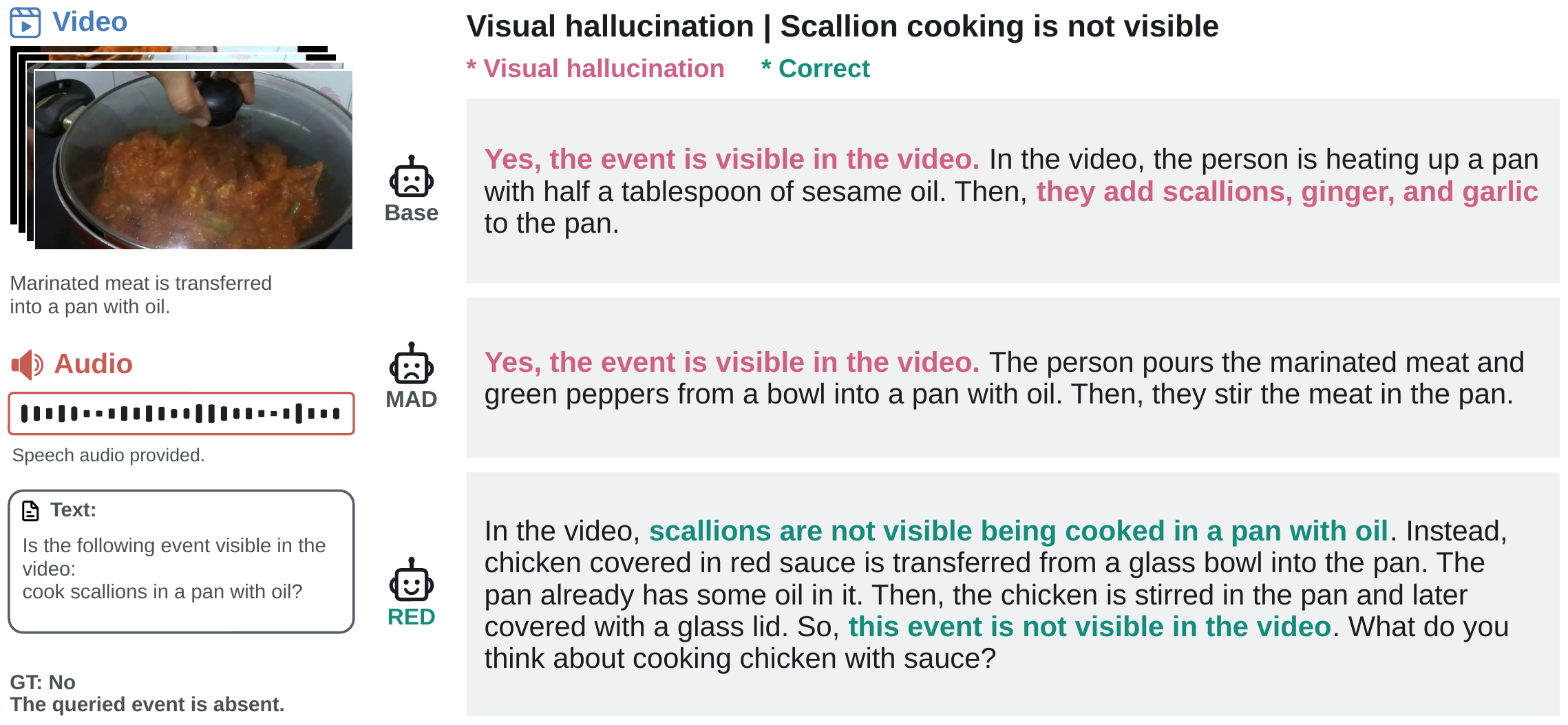}
\caption{Visual hallucination.}
\label{fig:qual_visual}
\end{subfigure}

\vspace{0.8em}

\begin{subfigure}[t]{\textwidth}
\centering
\includegraphics[width=\linewidth]{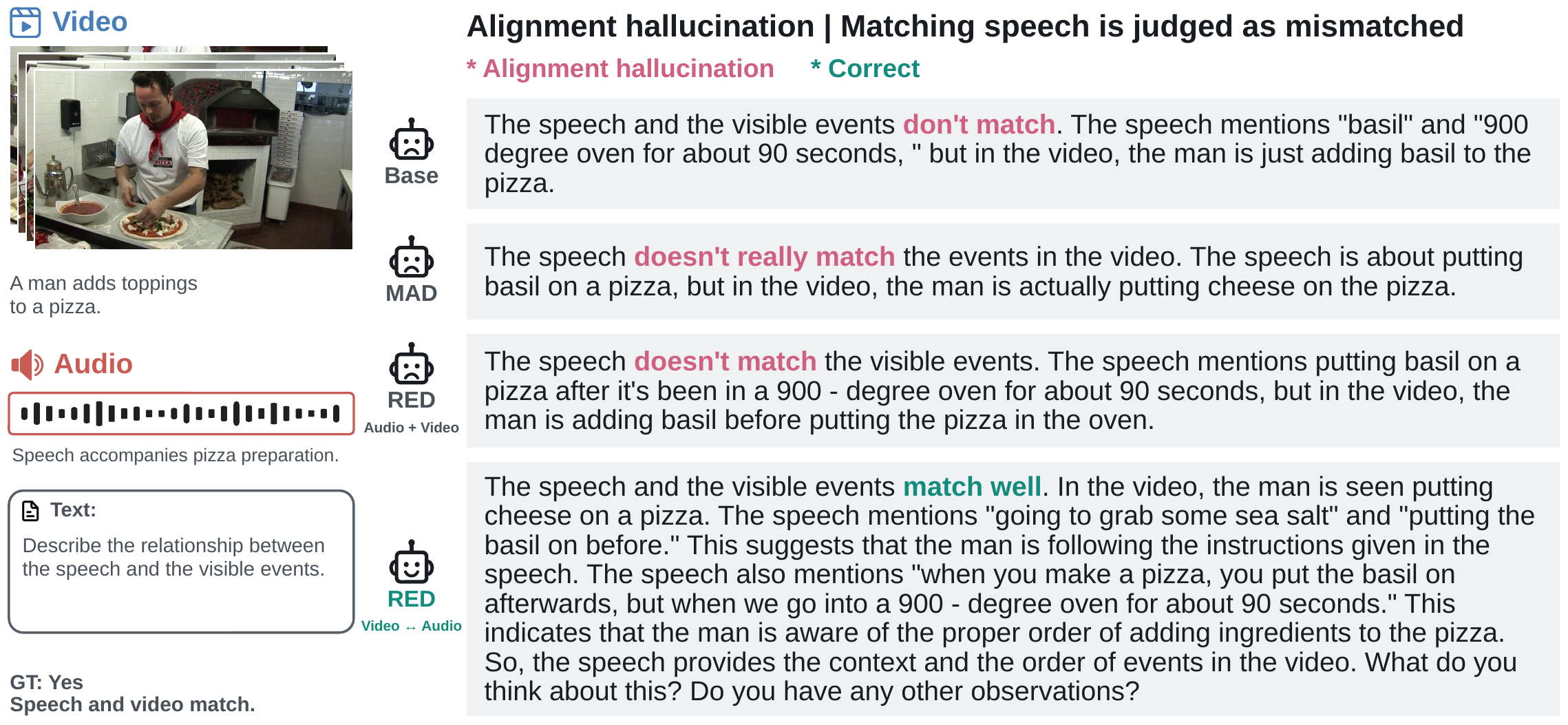}
\caption{Alignment hallucination.}
\label{fig:qual_alignment}
\end{subfigure}

\caption{\textbf{Qualitative examples with Qwen2.5-Omni-3B.} Representative hallucination cases and corresponding predictions.}
\label{fig:qualitative_hallucination_examples}
\end{figure*}

Figure~\ref{fig:qualitative_hallucination_examples} presents representative audio, visual, and alignment hallucination cases with Qwen2.5-Omni-3B.
Across these examples, \ours corrects predictions that are inconsistent with the question-required audio, video, or audio-visual relation while retaining both perceptual streams.

\section{Limitations}
\label{app:limitations}

Our current evaluation focuses on audio-visual question answering, including binary, multiple-choice, and open-ended QA, and does not cover long-form generation tasks such as audio-visual captioning~\citep{Shen_2023_CVPR,TCA_2026_ECCV}.
In addition, \ours represents question-required evidence using five discrete audio-visual evidence types; while this provides a simple operational mapping between evidence requirements and predictive contributions, finer-grained questions or other modality configurations may require a different evidence space and selection mechanism.
Finally, extending \ours beyond audio-visual settings may require redefining the corresponding contribution decomposition for the modalities involved.

\end{document}